\documentclass[journal]{IEEEtran}
\usepackage{algorithm}
\usepackage{algorithmic}
\usepackage{amsmath}
\usepackage{amssymb}
\usepackage{etex}
\usepackage{epsfig}
\usepackage{morefloats}

\newcommand{\argmin}{\operatornamewithlimits{argmin}}

\begin{document}

\title{NK Hybrid Genetic Algorithm for Clustering}

\author{Renato~Tin\'os,~\IEEEmembership{Member,~IEEE,}
        Liang~Zhao,~\IEEEmembership{Senior Member,~IEEE,}
        Francisco~Chicano,       
        and~Darrell~Whitley
\thanks{R. Tin\'os and L. Zhao are with the Department of Computing and Mathematics, University of S\~ao Paulo, Ribeir\~ao Preto, SP, Brazil e-mail: rtinos@ffclrp.usp.br , zhao@usp.br.}
\thanks{F. Chicano is with the Department of Lenguajes y Ciencias de la Computaci\'on, University of M\'alaga, M\'alaga, Spain e-mail: chicano@lcc.uma.es.}
\thanks{D. Whitley is with the Department of Computer Science, Colorado State University, Fort Collins, CO, USA e-mail: whitley@cs.colostate.edu.}
\thanks{Manuscript received XXXXX.}}

\markboth{NK Hybrid GA for Clustering,~Vol.~XX, No.~XX, Month~20XX}%
{Tin\'os \MakeLowercase{\textit{et al.}}: NK Hybrid GA for Clustering}

\maketitle

\begin{abstract}
The NK hybrid genetic algorithm for clustering is proposed in this paper. 
In order to evaluate the solutions, the hybrid algorithm uses the NK clustering validation criterion 2 (NKCV2). 
NKCV2 uses information about the disposition of $N$ small groups of objects. 
Each group is composed of $K+1$ objects of the dataset. 
Experimental results show that density-based regions can be identified by using NKCV2 with fixed small $K$. 
In NKCV2, the relationship between decision variables is known, which in turn allows us to apply gray box optimization. 
Mutation operators, a partition crossover, and a local search strategy are proposed, all using information about the relationship between decision variables. 
In partition crossover, the evaluation function is decomposed into $q$ independent components;
partition crossover then deterministically returns the best among $2^q$ possible offspring with computational complexity $O(N)$. 
The NK hybrid genetic algorithm allows the detection of clusters with arbitrary shapes and the automatic estimation of the number of clusters. 
In the experiments, the NK hybrid genetic algorithm produced very good results when compared to another genetic algorithm approach and to state-of-art clustering algorithms. 
\end{abstract}


%
\IEEEpeerreviewmaketitle

\section{Introduction}
\IEEEPARstart{C}{lustering} algorithms are important tools for the analysis and visualization of large complex datasets.  
Clustering is a grouping problem, where objects in the same clusters should be more similar to each other than to objects in other clusters.  
An important type of clustering is hard partitional clustering, where a single discrete label should be assigned for each object in the dataset \cite{halkidi2001}.  
Unsupervised learning algorithms are used to find good partitions in clustering.  
In this case, only internal validation criteria can be applied in clustering, i.e., the disposition of the objects in the input space is the only information available to evaluate partitions. 

Unfortunately, a unique definition for clusters does not exist \cite{kaufman2009} and several internal validation criteria have been proposed \cite{jaskowiak2015}. 
For example, the validation criterion used in k-means is the sum of distances from objects to the centroids of the clusters \cite{macqueen1967}. 
The validation criterion used in k-means is not appropriate when the number of clusters is unknown. 
Besides, among many of the internal validation criteria, it works well only in problems with spherical clusters \cite{rodriguez2014}. 

When employed in clustering and other grouping problems \cite{falkenauer1998}, evolutionary algorithms and other metaheuristics use fitness functions based on internal validation criteria \cite{hruschka2009}.   
In fact, external validation criteria could be used to evaluate the individuals; however, the information about the labels is not present in real-word clustering problems. 
Different internal validation criteria have been used to evaluate the partitions given by the individuals. 
For example, in \cite{tseng2001}, the weighted sum of cluster intradistances and interdistances is employed to evaluate solutions in a genetic algorithm used to group small clusters into larger clusters. 
In \cite{hruschka2003}, the silhouette width criterion is used in the clustering genetic algorithm. 
Both internal validation criteria, as well as others that are generally used to evaluate individuals in evolutionary algorithms, are indicated for datasets with spherical clusters. 

In \cite{tinos2016}, we proposed the \emph{NK clustering validation} (NKCV) criterion. 
In the NKCV criterion, neighbourhood relations inside small groups of objects are used in order to improve the identification of clusters with arbitrary shapes.  
The NKCV function is composed of $N$ subfunctions, where $N$ is the number of objects in the dataset. 
Each subfunction is influenced by a small group of $K+1$ objects. 
Other clustering internal criteria were proposed to allow the detection of clusters with arbitrary shapes \cite{moulavi2014}, but only a fitness function based on the NKCV criterion allows the application of partition crossover \cite{whitley09,tinos2015}. 
Partition crossover is an efficient deterministic recombination operator that finds the best of $2^{q}$ possible offspring, where $q$ is the number of recombining subsets, at $O(N)$ cost per recombination if $K$ is a constant.  
Partition crossover can only be applied when the evaluation function is decomposable, i.e., it cannot be applied when traditional internal clustering validation criteria are employed.

Here, we introduce the NK hybrid \emph{genetic algorithm} (GA) for clustering.
The NK hybrid GA evaluates candidate solutions using NKCV2, a new version of the NKCV criterion. 
Unlike other clustering validation criteria, NK clustering validation criteria make explicit the relationship between decision variables. 
As a consequence, it is possible to use gray box optimization \cite{whitley2016}. 
In other words, we can use the information about the relationship between variables, described in the evaluation function, to efficiently search for new solutions.  
In this work, we propose new search operators and strategies that enhance the optimization process in clustering by using the information about the relationship between variables. 
Gray box optimization is a new promising approach for clustering; similar search operators and strategies have shown to be very efficient in other applications \cite{tinos2015,chicano2016a,chicano2016b}.

The main contributions of the paper are: 
i) Proposing a new NK clustering validation criterion (NKCV2) that uses the density of data points, as well as the distances, to define the local groups. 
In the NKCV criterion \cite{tinos2016}, only the distances between the objects are used. 
Unlike the NKCV criterion, the NKCV2 criterion can be used in problems with noise. 
Using a definition similar to the one presented in \cite{moulavi2014}, the NKCV2 criterion considers that noisy objects present small local density and are distant from other objects in the group. 
ii) Proposing mutation operators and a local search method that use the information about the structure of an evaluation function given by the NKCV2 criterion. 
In other words, the operators explicitly explore the information about the interaction between decision variables obtained by analysing the subfunctions that form the evaluation function. 
This is also true for the partition crossover. 
The proposed operators and local search method can be applied to clustering only when the NKCV2 function is employed to evaluate the solutions. 
iii) Proposing the NK hybrid GA, that incorporates the NKCV2 function to evaluate the solutions, partition crossover, the new mutation operators, and the new local search method. 

The proposed NK hybrid GA allows to identify clusters with arbitrary shape and the automatic estimation of the number of clusters. 
Although algorithms based on the density of the points in the dataset, like DBSCAN (\emph{density-based spatial clustering of applications with noise}) \cite{ester1996} and the \emph{density peaks} (DP) clustering algorithm \cite{rodriguez2014}, also efficiently identify clusters with arbitrary shape and automatically estimate the number of clusters, those are sensitive to the algorithm parameters or to the manual selection of candidates for prototypes. 
In fact, it is possible to run the algorithms with different combinations of parameters and find the best partition according to an evaluation function based on an internal validation criterion. 
The same approach can be used to select the best parameter $k$, that indicates the number of clusters, in k-means. 
The automatic selection of the number of clusters by generating a large number of solutions and selecting the best ones according to the NKCV2 criterion is intrinsic to the proposed hybrid genetic algorithm. 

The remainder of the paper is organized as follows. 
Related work and some concepts related to hard partitional clustering are presented in Section~\ref{sec:cluster}. 
Sections \ref{sec:nkeval_func} and \ref{sec:GA} present the methodology proposed in this work: Section~\ref{sec:nkeval_func} presents the NKCV2 criterion, while Section~\ref{sec:GA} presents the NK hybrid genetic algorithm. 
Comparative experiments with k-means, DBSCAN, DP clustering algorithm, and another genetic algorithm approach are presented in Section~\ref{sec:exp}.
Finally, the paper is concluded in Section~\ref{sec:con}.

\section{Background and Related Work}
\label{sec:cluster}
Each object is represented by an $l$-dimensional attribute vector $\mathbf{y}_{j}$, while a dataset with $N$ objects is represented by $Y=\{\mathbf{y}_{1} \ldots \mathbf{y}_{N} \}$. 
A \emph{hard partition} is a collection $C= \{ C_1, \ldots , C_{N_c} \}$ with $N_c$ subsets (clusters) $C_i \neq \varnothing$, such that $C_1 \cup C_2 \cup \ldots \cup C_{N_c}=Y$, and $C_i \cap C_j = \varnothing$ for all $i \neq j$. 
In hard partitional clustering, given a dataset $Y$, the best hard partition should be found. 
Clustering is an NP-hard grouping problem \cite{falkenauer1998}. 

Three types of encoding are common in evolutionary algorithms applied to hard partitional clustering \cite{hruschka2009}. 
In the binary encoding, 
a bit is associated with each object. 
A value of 1 at the $i$-th position of the solution vector $\mathbf{x}$ means that the $i$-th object is a prototype for a cluster. 
A prototype is a particular vector representing a cluster. 
In many works, centroids and medoids of cluster distributions are used as prototypes. 
A different kind of binary representation uses a binary matrix to indicate if the $i$-th object (at the $i$-th column) belongs to the $j$-th cluster (at the $j$-th line).
On the other hand, the coordinates of each prototype are represented in the solution vector $\mathbf{x} \in \mathbb{R}^{l}$ in the real representation. 
In this case, the number of clusters, $N_c$, is known in advance. 

In the integer representation, used in the hybrid genetic algorithm proposed here, the number of clusters can change during the optimization process. 
A solution representing a partition is given by an $N$-dimensional vector $\mathbf{x} \in \mathbb{N}^N$. 
An element $x_j=i$ indicates that $\mathbf{y}_j \in C_i$, i.e., the object $\mathbf{y}_j$ belongs to cluster $C_i$. 
When noise is considered, label 0 can be used to indicate a noisy object, i.e., $x_i=0$ indicates that the $i$-th object is noise.
One can observe that the integer representation is redundant. 
For example, a partition can be represented by $\mathbf{x}=\{1,1,1,1,3,3,2,2,2,2\}$ or by $\mathbf{x}=\{2,2,2,2,1,1,3,3,3,3\}$. 
Such redundancy can cause problems when two solutions are recombined. 
A way to reduce the effects of redundancy is to use a renumbering process whenever two solutions are recombined \cite{falkenauer1998}.

The proposed NK hybrid GA (Section~\ref{sec:GA}) is compared in Section~\ref{sec:exp} to another GA approach for clustering: the \emph{clustering GA} (CGA) proposed in \cite{hruschka2003}. 
The CGA contains strategies and operators that are usually employed in population metaheuristics applied to clustering. 
In this way, it is a good option for testing the performance of our approach relative to other population metaheuristics. 
Like the NK hybrid GA, CGA uses integer codification. 
The silhouette width criterion (Section~\ref{sec:eval_func}) is employed to evaluate the individuals. 
Two mutation operators are used (\emph{merge} and \emph{split}). 
In the first, a randomly chosen cluster is merged with the nearest cluster. 
The distance between centroids is used to measure the distance between clusters. 
In the second, a random cluster is split: objects closer to the farthest object (from the centroid of the original cluster) are moved to the new cluster. 
CGA uses a context-sensitive recombination operator similar to the crossover used in the grouping GA proposed by Falkenauer \cite{falkenauer1998}. 
In the CGA crossover, $k_1$ clusters of the offspring are inherited from the first parent. 
The next step is to copy the non-overlapping clusters from the second parent. 
Finally, each of the remaining objects is allocated to the nearest cluster according to their distances to the centroids.    
Crossover is applied with rate $p_c$; if crossover is not applied, one of the two mutations are applied with equal probability.
\vspace{-3mm}
\subsection{Clustering Internal Validation}
\label{sec:eval_func}
Many clustering internal validation measures have been proposed for hard partitional clustering. 
The k-means algorithm uses the sum of square Euclidean distances from the objects to the centroids of the clusters, i.e.:
\vspace{-2mm}
\begin{equation}
\label{eq:eval_f1}
f(\mathbf{x}) = \sum_{i=1}^{N_c(\mathbf{x})} \sum_{\mathbf{y}_j \in C_i(\mathbf{x}) } \| \mathbf{y}_j - \mathbf{z}_i(\mathbf{x})  \| ^2
\end{equation}
where $\| . \|$ represents the Euclidean norm, $\mathbf{z}_i(\mathbf{x})$ it the $i$-th centroid, and $N_c(\mathbf{x})=k$ is the number of clusters in solution $\mathbf{x}$. 
Other distance measures can also be used.

Prototypes are necessary for computing Eq.~\eqref{eq:eval_f1}. 
Also, this validation criterion is not generally used in population metaheuristics approaches applied to clustering because the number of clusters ($k$) should be known in advance. 
Alternatively, the silhouette width criterion \cite{kaufman2009} can be employed:
\vspace{-2mm}
\begin{equation}
\label{eq:eval_f3}
f(\mathbf{x}) = \frac{1}{N} \sum_{j=1}^{N} s(\mathbf{y}_j,\mathbf{x})
\end{equation}
where:
\vspace{-2mm}
\begin{equation}
\label{eq:silhouette}
s(\mathbf{y}_j,\mathbf{x}) = \frac{b(\mathbf{y}_j,\mathbf{x})-a(\mathbf{y}_j,\mathbf{x})}{\max \{ b(\mathbf{y}_j,\mathbf{x}),a(\mathbf{y}_j,\mathbf{x}) \}}
\end{equation}
Let $d(\mathbf{y}_j,C_i(\mathbf{x}))$ be the average dissimilarity (e.g., the average Euclidean distance) of $\mathbf{y}_j$ to all objects belonging to cluster $C_i(\mathbf{x})$. 
Then, if $\mathbf{y}_j \in C_i(\mathbf{x})$,  $a(\mathbf{y}_j,\mathbf{x})=d(\mathbf{y}_j,C_i(\mathbf{x}))$. 
In other words, $a(\mathbf{y}_j,\mathbf{x})$ is the average dissimilarity of $\mathbf{y}_j \in C_i(\mathbf{x})$ to all objects in the same cluster $C_i(\mathbf{x})$. 
The term  $b(\mathbf{y}_j,\mathbf{x})$ is the average dissimilarity of $\mathbf{y}_j$ to its closest neighbour cluster, i.e.,  
\vspace{-2mm}
\begin{displaymath}
b(\mathbf{y}_j,\mathbf{x}) = \min _{i=1,\ldots,N_c(\mathbf{x}) ; \mathbf{y}_j \not\in C_i(\mathbf{x})} d(\mathbf{y}_j,C_i(\mathbf{x}))
\end{displaymath}
When applied as the evaluation function, Eq.~\eqref{eq:eval_f3} should be maximized. 
Also, by definition, $s(\mathbf{y}_j,\mathbf{x})=0$ if there is only one object in the cluster.  
The procedure to compute the evaluation function given by Eq.~\eqref{eq:eval_f3} has cost $O(N^2)$. 
However, this cost can be reduced to $O(N)$ if the distances of $\mathbf{y}_j$ to the centroids are used instead of the distances to all objects in the clusters \cite{hruschka2009}. 
The silhouette width criterion has been used as evaluation function in many population metaheuristics approaches applied to clustering \cite{hruschka2009}. 

Unlike the silhouette width criterion, the \emph{density-based clustering validation} (DBCV) \cite{moulavi2014} was proposed for the validation of clusters with arbitrary shapes. 
DBCV was developed for the validation of clusters composed of dense objects, separated by low dense areas. 
In order to capture the shape and density of clusters with arbitrary shape, a minimum spanning tree is built for each cluster. 
The idea of using the minimum spanning trees, and more generally, graphs, to represent the structure of the clusters is not new. 
It has already been used in other density-based validation criteria, e.g., in \cite{pal1997} and \cite{yang2004}.
The main contribution of DBCV is the use of a symmetric reachability distance to build the minimum spanning tree \cite{moulavi2014}. 
DBCV also incorporates a mechanism to handle noisy datasets.  
\vspace{-3mm}
\subsection{Local density}
In k-means \cite{macqueen1967}, Eq.~\eqref{eq:eval_f1} is minimized by successively recomputing the centroids of $k$ clusters and reallocating the objects to the closest current centroids. 
By allocating each object to the cluster with the nearest centroid, spherical clusters are found. 

Another approach is to use the local density of the objects in order to create the clusters. 
The local density of the $i$-th object of a dataset with $N$ objects is defined as: 
\vspace{-2mm}
\begin{equation}
\label{eq:density}
\rho_{i} = \sum_{j=1}^{N} \mathbf{K}(\mathbf{y}_i-\mathbf{y}_j)
\end{equation}
where $\mathbf{K}(\mathbf{y}_i-\mathbf{y}_j)$ is a kernel function. 
If the flat kernel is employed, $\mathbf{K}(\mathbf{y}_i-\mathbf{y}_j)=1$ if $\|\mathbf{y}_i-\mathbf{y}_j\|<\epsilon$ and 0 otherwise. 
In this case, $\rho_{i}$  is equal to the number of objects 
where the distance to $\mathbf{y}_i$ is smaller than the cutoff distance $\epsilon$. 
We use, in this work, the Gaussian kernel, given by:
\vspace{-2mm}
\begin{equation}
\label{eq:gaussian_kernel}
\mathbf{K}(\mathbf{y}_i-\mathbf{y}_j) = e^{\frac{-\|\mathbf{y}_i-\mathbf{y}_j\|^2}{2 \epsilon^2}}
\end{equation}

In DBSCAN \cite{ester1996}, the definition of the parameters has a great impact in the performance of the algorithm. 
Clusters are formed by connecting neighbouring objects if their densities ($\rho_i$) are higher than a threshold $minPts$. 
Otherwise, the object is considered as noise. 
The definition of the parameters has less importance in the DP clustering algorithm \cite{rodriguez2014}. 
Besides the local density $\rho_i$, the distance to the nearest object with higher density, $\delta_i$, is computed for each object $\mathbf{y}_i$. 
The next step is to plot $\rho_i \times \delta_i$ for all objects and to manually find the objects presenting the highest densities and the highest distances $\delta_i$. 
In other words, a rectangle is manually defined in the upper-right corner of the plot $\rho_i \times \delta_i$ in order to select the objects with highest densities and highest distances $\delta_i$. 
Those objects are considered as prototypes of the clusters, while the objects with lower densities and higher distances are considered as noise. 

The DP clustering algorithm involves a manual selection of the number of prototypes. 
However, it is not difficult to think of a mechanism similar to the one used to select $k$ in k-means, i.e., by 
running the algorithm with different parameters and using internal validation to select the best partitions. 
In fact, we use this approach in Section~\ref{sec:exp} when the DP algorithm is employed. 
A more serious disadvantage is the application in problems where the points of highest densities are not the prototypes of the clusters. 
For example, one can imagine circular clusters where all the objects have the same local density.   

Both DP clustering algorithm and DBSCAN can be applied in problems where the number of clusters is unknown beforehand.  
Also, both algorithms have intrinsic mechanisms to deal with noisy objects. 
Like other density-based clustering algorithms, e.g., the mean shift algorithm \cite{cheng1995}, they are able to detect clusters with arbitrary forms. 

Here, the local density of the objects (Eq.~\eqref{eq:density}) is used to build the interaction graph for the NK clustering internal validation criterion. 
We use the rule of thumb proposed in \cite{rodriguez2014} for the definition of the cutoff distance: $\epsilon$ is chosen in order that the average number of neighbours is $2\%$ of the total number of objects in the dataset. 
This rule showed to be robust to several datasets. 
More accurate ways to estimate $\epsilon$ exist and can be used for datasets with few objects \cite{cheng1995}.

\section{NK Clustering Internal Validation Criterion}
\label{sec:nkeval_func}
The silhouette width criterion (Eq.~\eqref{eq:eval_f3}) is composed of a sum of $N$ subfunctions. 
The same is done in the NKCV2 criterion. 
However, in the latter, each subfunction depends on a small number of objects $K<<N$, while each subfunction can depend on a large number of objects in the former. 
This is the main innovation of the NKCV2 criterion: using the local information in order to build the evaluation function. 
Neighbourhood relations inside a small subset of objects are used to compute the subfunctions. 
As the subsets are not necessarily disjoint, links are formed. 
The neighbourhood relations represent the linkage between elements of the solution vector. 

The NKCV2 is given by:
\vspace{-2mm}
\begin{equation} 
\label{eq:NKeval_f}
f(\mathbf{x}) = \sum_{i=1}^{N} f_i(\mathbf{x})
\end{equation}
where each subfunction $f_{i}: \mathbb{N}^{K+1} \rightarrow \mathbb{R}$ is influenced by a subset of $K+1$ elements of $\mathbf{x}$, i.e., the subfunctions $f_i$ are (K+1)-bounded. 
A mask $\mathbf{m}_i \in \mathbb{B}^N$ with $K+1$ ones can be used to represent the elements of $\mathbf{x}$ that influence subfunction $f_i$. 
If the element of $\mathbf{m}_i$ at position $j$ is equal to one, then $x_j$ influences subfunction $f_i$. 
As we will see in Section~\ref{sec:nk_subf}, the disposition of the objects in the space also 
influences each subfunction $f_i$. 
However, given a dataset, the disposition is fixed and we can consider that only the labels influence the subfunctions when a partition is evaluated. 

The interactions among the elements of $\mathbf{x}$ can be represented by an adjacency matrix $\mathbf{M} = [\mathbf{m}_1 \mathbf{m}_2 \ldots \mathbf{m}_N]$. 
The directed graph with adjacency matrix $\mathbf{M}$ is called \emph{interaction graph} and is represented by $G_{ep}$. 
In the interaction graph, the vertex $v_i$, $i=1,\ldots, N$, represents the $i$-th element of the solution vector $\mathbf{x}$. 
The edge $(v_j,v_i)$ indicates that element $x_j$ influences $f_i$.

The term \emph{group} is used here for the subset of $K+1$ objects interacting in a subfunction. 
We expect that groups are uniform, i.e., are in the same cluster, if their respective objects are close. 
However, groups should not be uniform if they are composed of distant objects. 
By using Eq.~\eqref{eq:NKeval_f}, clusters with spherical distributions will not always be favoured. 
The distributions of labels inside the groups are important for the evaluation of the partition. 
Decreasing $K$, smaller groups are considered.

One can observe that the NKCV2 function (Eq.~\eqref{eq:NKeval_f}) is similar to the function used in the NK landscape model \cite{kauffman93}. 
In the NK landscape model, random or adjacent interaction graphs are generally employed and the values for the subfunctions $f_i$ are randomly generated.
In NKCV2, the interaction graph is given by the disposition of the objects in the attribute space. 
Therefore, it does not depend on the partitions ($\mathbf{x}$). 
The values of the subfunctions $f_i$ are not random; they depend on the disposition of the objects in the attribute space and also on the partitions ($\mathbf{x}$). 
The method for building the interaction graph, i.e., to define the masks $\mathbf{m}_i$, is described in Section~\ref{sec:nk_epgraph}.
The subfunctions $f_i$ are described in Section~\ref{sec:nk_subf}.
\vspace{-3mm}
\subsection{Building the interaction graph}
\label{sec:nk_epgraph}
After creating a vertex with self-loop, $v_i$, for each object $\mathbf{y}_i$ of the dataset, two procedures are applied to create the interaction graph $G_{ep}$. 
In the first procedure, for each vertex $v_i$ representing $\mathbf{y}_i$, the vertex representing the nearest object with higher density, $v_{a_i}$, is identified. 
Then, an edge is created from $v_{a_i}$ to $v_{i}$.  
We expect that, by connecting objects with densities that are local maxima, connections between different clusters can be formed. 
This is important because objects in a cluster should be more similar to other objects in the same cluster than to objects in other clusters. 
Computing the densities of the objects involves computing all the distances between the objects, what has $O(N^2)$ time complexity. 
However, computing the densities involves also computing the cutoff distance $\epsilon$  used in Eq.~\eqref{eq:gaussian_kernel}, what has $O(N^2 \log N)$ time and memory complexity when the method presented in \cite{rodriguez2014} is used. 

In the second procedure, edges between vertices representing close objects are added. 
Incident edges from the nearest objects are added to each vertex $v_i$ until its indegree becomes equal to $K+1$. 
The time complexity for the second procedure is $O(N^2)$. 
 
\begin{figure*}[ht]
\centering
\includegraphics[width=0.80\textwidth]{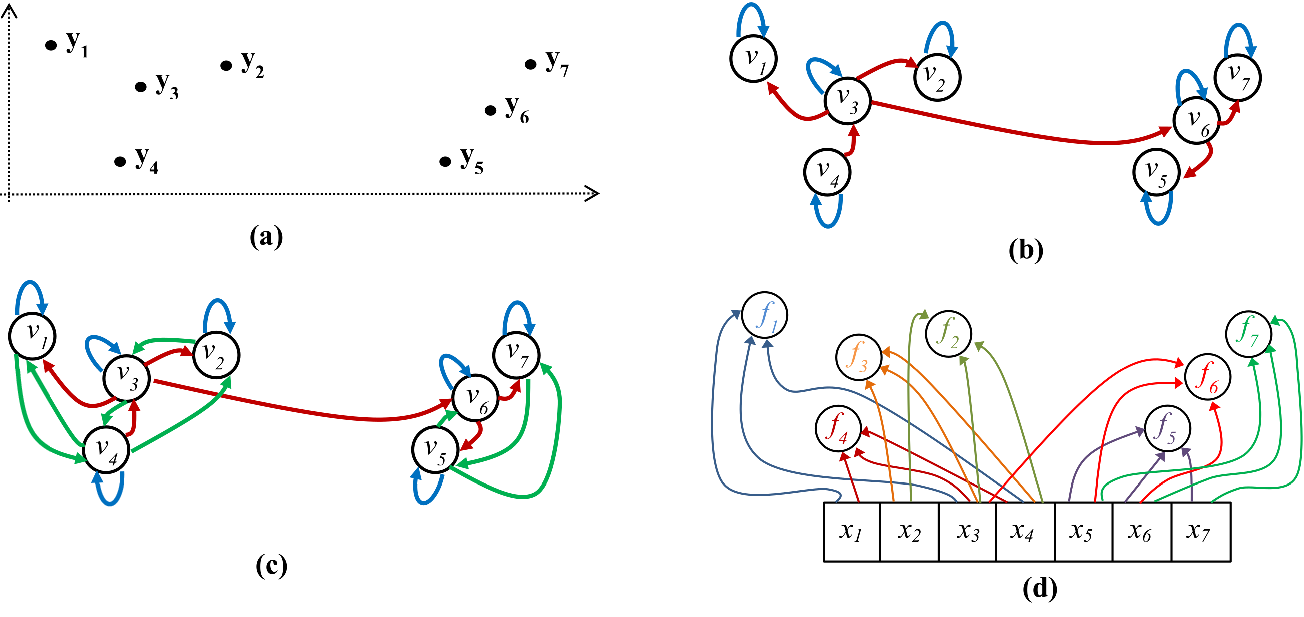}
\caption{Building the interaction graph. 
a) Dataset with 7 objects ($N=7$). 
b) Each object of the dataset is associated with a vertex with self-loop. 
For each vertex, an incident edge from the vertex representing the nearest object with higher density is created (red edges). 
By definition, the incident edge for the vertex with highest density is from the vertex representing the nearest object. 
c) The next step is to add edges between near objects until each vertex has indegree equal to $K+1$ (in this example, $K=2$). 
The interaction graph has $N=7$ vertices and $N(K+1)$ edges. 
d) Each subfunction $f_i$ is influenced by a group of $K+1$ decision variables. 
The group of labels influencing subfunction $f_i$ is defined by the $K+1$ incident edges to $v_i$.}
\vspace{-3mm}
\label{fig:ex_epgraph}
\end{figure*}

An example for building $G_{ep}$ is shown in Figure \ref{fig:ex_epgraph}. 
The evaluation function (Eq.~\eqref{eq:NKeval_f}) created from the interaction graph (Figure \ref{fig:ex_epgraph}.c) is:
\vspace{-2mm}
\begin{eqnarray} 
	 f(\mathbf{x}) 	& = & f_1(x_1,x_3,x_4)+f_2(x_2,x_3,x_4)+f_3(x_2,x_3,x_4) \nonumber \\ \vspace{-3mm}
					&	& +f_4(x_1,x_3,x_4)+f_5(x_5,x_6,x_7)\nonumber \\ \vspace{-3mm}
					& 	& +f_6(x_3,x_5,x_6)+f_7(x_5,x_6,x_7)					
\end{eqnarray}

\vspace{-3mm}
\subsection{Subfunctions $f_i$}
\label{sec:nk_subf}
The subfunctions $f_i(.)$ are influenced by the labels ($\mathbf{x}$) and disposition of the objects inside the groups ($\mathbf{y}$). 
For simplicity, the dependency on $\mathbf{y}$ is not shown in $f_i(.)$. 
The equation for subfunctions $f_i(\mathbf{x})$ is defined as follows:
\vspace{-2mm}
\begin{equation} 
\label{eq:f_i}
f_i(\mathbf{x}) = \sum_{j \in M_i , j \neq i} \alpha(x_i,x_j)
\end{equation}
where $M_i$ indicates the indices of the $K+1$ objects inside the $i$-th group, i.e., all indices $j$ for the vertices $v_j$ such that $(v_j,v_i) \in G_{ep}$. 
The subset $M_i$ contains the index $i$ and other $K$ indices of the objects that influence function $f_i$. 
In Eq.~\eqref{eq:f_i}, the term $\alpha(x_i,x_j)$ is given by:
\vspace{-2mm}
\begin{equation} 
\label{eq:alpha_i}
\alpha(x_i,x_j)= \left\{ 
\begin{array}{ll}
\alpha_{noise} & \textrm{if $x_i = 0$}\\
\alpha_{in} & \textrm{if $x_i = x_j$}\\
\alpha_{out} & \textrm{otherwise} \\
\end{array} \right.
\end{equation}

\begin{figure*}[ht]
\centering
\includegraphics[height=3.0cm, width=0.35\textwidth]{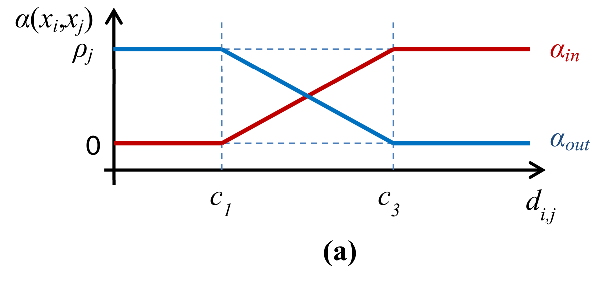}
\includegraphics[height=3.0cm, width=0.35\textwidth]{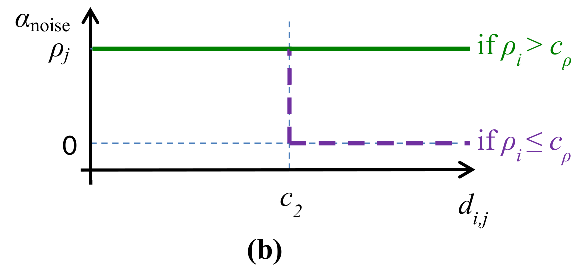}
\caption{Subfunctions $f_i$. }
\label{fig:ex_fi}
\vspace{-3mm}
\end{figure*}

When $x_i = 0$, $\alpha(x_i,x_j)$ is given by:
\vspace{-2mm}
\begin{equation} 
\label{eq:alpha_noise}
\alpha_{noise} = \left\{ 
\begin{array}{ll}
0 & \textrm{if $d_{i,j} > c_{2}$ and $\rho_i \leq c_{\rho}$}\\
\rho_j & \textrm{otherwise}\\
\end{array} \right.
\end{equation}
where $d_{i,j}=\| \mathbf{y}_i-\mathbf{y}_j \|$ is the distance between objects $\mathbf{y}_i$ and $\mathbf{y}_j$, and $c_{2}, c_{\rho}\in \mathbb{R}^+$ are thresholds (see Figure \ref{fig:ex_fi}.b). 
We are using the same definition employed in \cite{rodriguez2014} for noise, i.e., the $i$-th object is considered noise if its density ($\rho_i$) is small and if it is distant from other objects in the group. 

When the labels $x_i$ and $x_j$ for the two objects inside group $M_i$ are equal, we expect that the objects are close. Thus:
\vspace{-2mm}
\begin{equation} 
\label{eq:alpha_in}
\alpha_{in} = \left\{ 
\begin{array}{ll}
0 & \textrm{if $d_{i,j} < c_{1}$}\\
\frac{d_{i,j} - c_{1}}{c_{3} - c_{1}}\rho_j & \textrm{if $c_{1} \leq  d_{i,j} \leq c_{3}$}\\
\rho_j & \textrm{if $d_{i,j} > c_{3}$}\\
\end{array} \right.
\end{equation}
where $c_1,c_3 \in \mathbb{R}^+$ are thresholds (see Figure \ref{fig:ex_fi}.a). 
 
Otherwise, i.e., when the labels for the two objects $x_i$ and $x_j$ inside group $M_i$ are different, we expect that the objects are distant: 
\vspace{-2mm}
\begin{equation} 
\label{eq:alpha_out}
\alpha_{out} = \left\{ 
\begin{array}{ll}
\rho_j & \textrm{if $d_{i,j} < c_{1}$}\\
\frac{c_{3}-d_{i,j}}{c_{3} - c_{1}}\rho_j  & \textrm{if $c_{1} \leq  d_{i,j} \leq c_{3}$}\\
0 & \textrm{if $d_{i,j} > c_{3}$}\\
\end{array} \right.
\end{equation}
One can observe that the density of the $j$-th object, $\rho_j$, is employed in the equations. 
Objects with higher density have more impact on $f_i$ than objects with lower density. 

The threshold $c_{\rho}$ is computed based on the mean ($m_{\rho}$) and standard deviation ($\sigma_{\rho}$) of the densities $\rho_i$, for $i=1,\ldots,N$. 
The thresholds $c_1$, $c_2$, and $c_3$ are computed based on the mean ($m_{y}$) and standard deviation ($\sigma_{y}$) of the distances $d_{i,j}$ inside all groups.
The equations for computing the thresholds are based on the 3 sigma rule (or 68-85-99.7 rule). 
The equations are:
\vspace{-2mm}
\begin{equation} 
\label{eq:c_1}
c_1 = m_{y}
\end{equation}
\vspace{-3mm}
\begin{equation} 
\label{eq:c_2}
c_2 = m_{y}+\sigma_{y}
\end{equation}
\vspace{-3mm}
\begin{equation} 
\label{eq:c_3}
c_3 = m_{y}+2 \sigma_{y}
\end{equation}
\vspace{-3mm}
\begin{equation} 
\label{eq:c_rho}
c_{\rho} = \left\{ 
\begin{array}{ll}
m_{\rho}-\sigma_{\rho} & \textrm{if $m_{\rho}-\sigma_{\rho}>0$}\\
\frac{m_{\rho}}{2} & \textrm{otherwise}\\
\end{array} \right.
\end{equation}
 
If the matrix of distances between the objects is available, the procedure to compute $c_1$, $c_2$, and $c_3$ has $O(N K)$ time complexity. 
If the densities $\rho_i$ are available, the procedure to compute $c_{\rho}$ has $O(N)$ time complexity. 
The thresholds can be finely adjusted for each dataset. 
However, using general rules based on the means and standard deviations for computing $c_1$, $c_2$, $c_3$, and $c_{\rho}$ is much simpler. 
Experiments (not shown here) indicate that the NKCV2 criterion is robust to the choice of the thresholds. 
Also, experimental results (Section~\ref{sec:exp}) indicate that setting the thresholds based on equations (\ref{eq:c_1})-(\ref{eq:c_rho}) results in good performance for diverse datasets.

Finding the interaction graph $G_{ep}$ and computing the thresholds and densities are done only one time for each dataset. 
The labels of the objects, given in $\mathbf{x}$, are not used for building $G_{ep}$ and computing the thresholds and densities. 
If more than one partition is evaluated for a given clustering instance, it is not necessary to build 
$G_{ep}$ again and recompute the thresholds and densities. 
This is the case for meta-heuristics applied to clustering, because several individuals are evaluated for only one dataset. 
Given $G_{ep}$ as well as the thresholds and densities, the time and memory complexity for evaluating one individual using Eq.~\eqref{eq:NKeval_f} is $O(N K)$, that is $O(N)$ if $K$ is a constant.

\section{NK Hybrid Genetic Algorithm}
\label{sec:GA}
In the NK hybrid GA, individuals represent clustering partitions using the integer encoding (Section~\ref{sec:cluster}). 
The individuals are evaluated using the NKCV2 function (Eq.~\ref{eq:NKeval_f}), that is similar to the function used in the NK landscape model.  
The NK landscape model belongs to a class of optimization problems where information about the relationship between variables is available by the inspection of the evaluation function \cite{whitley2016}. 
In this class of problems, the interaction graph, $G_{ep}$, can be obtained by analysing the problem structure, i.e., the variables that interact in the subfunctions.  
Unlike black box optimization algorithms, gray box optimization algorithms use the problem structure, described in the evaluation function, for designing operators and strategies. 

The information about the interaction between variables in subfunctions of the NKCV2 function is employed in a mutation operator and in the local search proposed in this paper for clustering. 
Two other mutation operators that explore the local density of the objects are still proposed. 
The local search and the mutation operators are described in Section~\ref{sec:mut}. 
Partition crossover for clustering, presented in Section~\ref{sec:PX}, also uses the interaction graph. 
By using $G_{ep}$ and the common elements found in the parents, it is possible to decompose the evaluation function in order to find an efficient crossover mask.

The pseudo-code for the NK hybrid GA is presented in Algorithm~\ref{alg:hga}. 
The population $P$ is composed of $size_{pop}$ individuals. 
The initial population is formed by random individuals optimized by local search. 
Elitism and tournament selection are employed. 
Partition crossover is applied with rate $p_c$. 
Before applying crossover, the labels in one of the parents are mapped to the labels of the other parent using a renumbering process \cite{hruschka2009}. 
After crossover, clusters with same label are relabeled using procedure $fixLabels$. 
In procedure $fixLabels$, the labels of all clusters are compared. 
A cluster is relabeled whenever its label is already used by another cluster. 

When crossover is not applied, the individual is transformed using one of three types of mutation. 
The first type, that changes only one element of the parent solution, is applied in 60\% of the cases. 
Each of the other two types occur in 20\% of the mutations.
In the experiments with the NK hybrid GA (Section~\ref{sec:exp}), we used standard values for the parameters or values adjusted in previous experiments (not shown here). 
Our experience indicated that the performance of the GA is, in general, robust to the choice of parameters like crossover and mutation rates. 
In order to preserve the diversity of the population, 30\% of the population is replaced by random individuals optimized by local search every 100 generations. 
Local search is also applied in the remaining individuals. 
 
\begin{algorithm}[ht]
\footnotesize 
\caption{NK Hybrid Genetic Algorithm}
\label{alg:hga}
\begin{algorithmic}[1]
\FOR {$i$=1 \TO $size_{pop}$}
\STATE $\mathbf{x}$=randomIndividual()
\STATE $P(i)$=localSearch($\mathbf{x}$)
\ENDFOR
\STATE $t=1$
\WHILE {not (stopping criterion)}
\STATE $Q(1)$=selectBestSolution($P$)
\FOR {$i$=2 \TO $size_{pop}$}
\STATE $p_1$=selection($P$)	
\IF {rand([0,1]) $< p_{c}$}
\STATE $p_2$=selection($P$)	
\STATE $\mathbf{x}^{m}_{p_2} $= renumbering ($\mathbf{x}_{p_1}$, $\mathbf{x}_{p_2}$)
\STATE $Q(i)$=PX($\mathbf{x}_{p_1}$, $\mathbf{x}^{m}_{p_2}$)
\STATE $Q(i)$=fixLabels$\big(Q(i)\big)$
\ELSE
\STATE r=rand([0,1])
\IF {$r < 0.6$}
\STATE $Q(i)$=mutationNK($\mathbf{x}_{p_1}$)
\ELSIF {$r < 0.8$}
\STATE $Q(i)$=mutationMerge($\mathbf{x}_{p_1}$)
\ELSE 
\STATE $Q(i)$=mutationSplit($\mathbf{x}_{p_1}$)
\ENDIF
\ENDIF
\ENDFOR
\IF {$t$ mod $100 = 0$}
\STATE $Q(1)$=selectBestSolution($Q$)
\FOR {$i$=1 \TO $\lceil 0.7*size_{pop} \rceil$}
\STATE  $Q(i)$=localSearch \big($Q(i)$\big)
\ENDFOR
\FOR {$i=\lceil 0.7*size_{pop} \rceil+1$ \TO $size_{pop}$}
\STATE $\mathbf{x}$=randomIndividual()
\STATE  $Q(i)$=localSearch \big($\mathbf{x}$\big)
\ENDFOR
\ENDIF
\STATE $P=Q$;
\STATE $t=t+1$
\ENDWHILE
\end{algorithmic}
\end{algorithm}

\vspace{-3mm}
\subsection{Designing Mutation Operators and a Local Search Method Based on the NK Clustering Validation Criterion}
\label{sec:mut}
The methods proposed here use the interaction graph, $G_{ep}$, in order to change a solution $\mathbf{x}$. 
All methods use $\Delta$-evaluation of $f$ to speedup the computations. 
$\Delta_{i,v}(\mathbf{x})$ denotes the change in the value of $f(\mathbf{x})$ when variable $x_i$ is assigned a new value $v$. In formal terms, $\Delta_{i,v}(\mathbf{x}) = f(\mathbf{y}) - f(\mathbf{x})$, where $y_j=x_j$ for all $j\neq i$ and $y_i=v$.
In the $\Delta$-evaluation, it is necessary to recompute only the subfunctions influenced by the label of the $i$-th object.  
In the algorithms, $M_p$ is the subset of indices for the $K+1$ objects inside the $p$-th group, i.e., the neighbours with edges to $v_p$ in $G_{ep}$. 
The time complexity of computing $\Delta_{i,v}(\mathbf{x})$ is $O(K_{out} K)$, where $K_{out}$ is the maximum outdegree in $G_{ep}$. 

In the first mutation ($mutationNK$), the label of the $i$-th object, $x_i$, is changed to the label of one of the objects with index belonging to the $i$-th group (Algorithm \ref{alg:mutNK}). 
In other words, $x_i$ assumes the value of one of the $K$ variables $x_{j \neq i}$ that influence $f_i$. 
The chosen label is the one that generates the minimum value of $\Delta_{i,v}(\mathbf{x})$.  
The time complexity for finding the best solution for all  possible changes in $x_i$ is $O(K_{out} K^2)$. 

\begin{algorithm}[ht]
\footnotesize 
\caption{$\mathbf{x}'$=mutationNK($\mathbf{x}$)}
\label{alg:mutNK}
\begin{algorithmic}[1]
\STATE $\mathbf{x}'=\mathbf{x}$
\STATE $i=\textrm{randint}(1,\ldots,N)$
\STATE $k$ = $\argmin_{j \in M_i, j\neq i} (\Delta_{i,x'_j}(\mathbf{x}))$
\STATE $x'_i=x'_k$
\end{algorithmic}
\end{algorithm}

The other two mutation operators are cluster-oriented, i.e., they are not designed to change a single object, but to transform a whole cluster. 
In $mutationMerge$ (Algorithm~\ref{alg:mutMerge}), two clusters are merged, i.e., the objects in one cluster assume the labels of the other cluster. 
In $mutationSplit$ (Algorithm~\ref{alg:mutSplit}), one cluster is split, i.e., some objects of the selected cluster receives a new label. 
Several works in the literature use merge and split operators \cite{hruschka2009}. 
Here, we propose the use of the local densities of the objects when splitting and merging clusters. 
In Algorithms \ref{alg:mutMerge} and \ref{alg:mutSplit}, the subset $labels(\mathbf{x})$ contain $N_c(\mathbf{x})$ labels found in partition $\mathbf{x}$.

In $mutationMerge$, the label of one cluster ($cluster_1$) is randomly chosen. 
Then, the prototype of each cluster is found. 
We consider here that the prototype of a cluster is the object with highest density in the cluster. 
The next step is to choose the label of a second cluster ($cluster_2$) according to probabilities proportional to the scaled distances between the prototypes. 
Clusters close to $C_{cluster_1}$ have higher probability of being selected than distant clusters.  
The last step is to change the labels of all objects in $C_{cluster_2}$ to label $cluster_1$. 
The local densities and distances between objects are computed only one time for each dataset, i.e., they are computed only in the beginning of the run (Section~\ref{sec:nkeval_func}). 
Thus, the time complexity of Algorithm \ref{alg:mutMerge} is $O(K_{out} K |C_{cluster_2}|)$. 
Observe that the cost can be worse than the cost $O(NK)$ of evaluating a new solution using the NK evaluation function (Eq.~\eqref{eq:NKeval_f}) for large clusters, i.e., if $K_{out} | C_{cluster_2}|$ is close to $N$. 
This can occur because some subfunctions are evaluated more than one time when the $\Delta$-evaluation is done in Algorithm \ref{alg:mutMerge}.  
As a consequence, when $K_{out} |C_{cluster_2}|$ is close to $N$, it is better to evaluate the solution outside $mutationMerge$ using Eq.~\eqref{eq:NKeval_f}. 

\begin{algorithm}[ht]
\footnotesize 
\caption{$\mathbf{x}'$=mutationMerge($\mathbf{x}$)}
\label{alg:mutMerge}
\begin{algorithmic}[1]
\STATE $\mathbf{x}'=\mathbf{x}$
\STATE $cluster_1=\textrm{randint}(labels(\mathbf{x}))$
\STATE $(\rho_{m_{cluster_1}}, m_{cluster_1})= \max\limits_{j \in C_{cluster_1}(\mathbf{x})} (\rho_j)$
\FOR {$\forall i \in labels(\mathbf{x}), i \neq cluster_1$}
\STATE $(\rho_{m_i}, m_i)= \max\limits_{j \in C_i(\mathbf{x})} (\rho_j)$
\STATE $d_{m_i, m_{cluster_1}}^{-1} = \frac{1}{\| \mathbf{y}_{m_i} - \mathbf{y}_{m_{cluster_1}} \|}$
\ENDFOR
\STATE $s = \sum\limits_{i \in labels(\mathbf{x}), i \neq cluster_1} d_{m_i, m_{cluster_1}}^{-1} $
\STATE randomly choose $cluster_2$ using probabilities $p_{m_i, m_{cluster_1}}=\frac{ d_{m_i, m_{cluster_1}}^{-1} }{s}$, $\forall i \in labels(\mathbf{x}), i \neq cluster_1$ 
\FOR {$\forall i \in C_{cluster_2}(\mathbf{x})$}
\STATE $x'_i=cluster_1$
\ENDFOR
\end{algorithmic}
\end{algorithm}

In $mutationSplit$, the label of one cluster is randomly chosen according to the distribution of clusters' sizes. 
Larger clusters have higher probabilities of being split. 
The two objects with highest densities are then selected; they are the new prototypes for the two new clusters. 
The labels of the objects inside the clusters are defined according to their distances to the prototypes: if the $i$-th object is closer to the second prototype, than to the first prototype, then $x_i$ receives the new label. 
The time complexity of Algorithm~\ref{alg:mutSplit} is $O(K_{out} |C_{cluster_1}| K)$. 
Thus, like in $mutationMerge$, if $K_{out} |C_{cluster_1}|$ is close to $N$, it is better to evaluate the solution outside $mutationSplit$ using Eq.~\eqref{eq:NKeval_f}. 

\begin{algorithm}[ht]
\footnotesize 
\caption{$\mathbf{x}'$=mutationSplit($\mathbf{x}$)}
\label{alg:mutSplit}
\begin{algorithmic}[1]
\STATE $\mathbf{x}'=\mathbf{x}$
\STATE $s = \sum\limits_{i \in labels(\mathbf{x})} |C_i(\mathbf{x})| $
\FOR {$\forall i \in labels(\mathbf{x})$}
\STATE $p_{i}=\frac{|C_i(\mathbf{x})|}{s} $
\ENDFOR
\STATE randomly choose $cluster_1$ according to probabilities $p_{i}$, $\forall i \in labels(\mathbf{x})$
\STATE $(\rho_{m^1_{cluster_1}}, m^1_{cluster_1})= \max\limits_{j \in C_{cluster_1}(\mathbf{x})} (\rho_j)$
\STATE $(\rho_{m^2_{cluster_1}}, m^2_{cluster_1})= \max\limits_{j \in C_{cluster_1}(\mathbf{x}), j \neq m^1_{cluster_1} } (\rho_j)$
\STATE $label^{new}=\min\limits_{j \not\in labels(\mathbf{x}), j>0 } (j)$
\FOR {$\forall i \in C_{cluster_1}(\mathbf{x})$}
\IF {$\| \mathbf{y}_{i} - \mathbf{y}_{m^1_{cluster_1}} \|>\| \mathbf{y}_{i} - \mathbf{y}_{m^2_{cluster_1}} \|$}
\STATE $x'_i=label^{new}$
\ENDIF
\ENDFOR
\end{algorithmic}
\end{algorithm}

\begin{algorithm}[ht]
\footnotesize 
\caption{$\mathbf{x}'$=localSearch($\mathbf{x}$)}
\label{alg:ls}
\begin{algorithmic}[1]
\STATE $\mathbf{x}'=\mathbf{x}$
\WHILE {not (stopping criterion)}
\STATE $i=\textrm{randint}(1,\ldots,N)$
\STATE $k=\argmin_{j \in M_i} (\Delta_{i,x'_j}(\mathbf{x}))$
\IF {$\Delta_{i,x'_k}(\mathbf{x}) \leq 0$}
\STATE $x'_i=x'_k$
\ENDIF
\ENDWHILE
\end{algorithmic}
\end{algorithm}

Finally, we propose a local search method based on $mutationNK$. 
However, instead of computing the best improvement for each variable $x_i$, we use the first improvement strategy. 
Also, neutral moves, i.e., when $\Delta_{i,v}(\mathbf{x})=0$, are allowed. 
In Algorithm \ref{alg:ls}, the index $i$ for the variable to be changed is randomly chosen. 
Like in $mutationNK$, variable $x_i$ is changed to the label of one of the objects that influences $f_i$. 
Only the subfunctions affected by changing $x_i$ are recomputed when the new solution is evaluated. 
This procedure is repeated until stopping criterion is met. 

\vspace{-3mm}
\subsection{Partition Crossover for Clustering}
\label{sec:PX}
In \cite{whitley09}, Whitley et al. proposed a deterministic recombination operator, named \emph{partition crossover} (PX), for the travelling salesman problem. 
PX explores the decomposition of the evaluation function by finding the connected components of the weighted graph formed by the union of two parent solutions. 
If $q$ connected components are found, $2^q$ ways to recombine the solutions are possible by individually selecting the edges inside the connected components from one or another parent. 
In PX, it is possible to compute the partial evaluations of each connected component for each parent. 
Thus, if the best partial solutions are chosen, the best among $2^q$ possible offspring is found at $O(N)$ computational cost. 

In \cite{tinos2015}, Tin\'os et al. extended PX for K-bounded pseudo-Boolean optimization. 
An example of K-bounded pseudo-Boolean optimization is the NK landscape model \cite{kauffman93}. 
The key idea for PX in this case is the observation that, if both parents share the same value for one element of the vector solution, the influence of inheriting this element from one or another parent in subfunctions are equal. 
Thus, the edges from vertices with decision variables shared by both parents can be removed from the interaction graph $G_{ep}$. 
Finding the connected components in the resulting graph, 
named the recombination graph ($G_{rec}$), allows the decomposition of the evaluation function. 

In PX, the values of the subfunctions in the offspring are inherited from the parents. 
As a consequence, it is possible to write the evaluation function of a solution $\mathbf{x}'$, obtained by applying PX in two parents $\mathbf{x}^{p1}$ and $\mathbf{x}^{p2}$, as:
\vspace{-2mm}
\begin{equation} 
\label{eq:f_p_xy}
f(\mathbf{x}') = \sum_{S_j \in S^{p1}} h_{S_j}(\mathbf{x}^{p1}) + \sum_{S_j \in S^{p2}} h_{S_j}(\mathbf{x}^{p2})
\end{equation}
where $S_j$ indicates the subset of variables at the $j$-th connected component and $S =S^{p1} \cup S^{p2} $ indicates that the set of indexes of all component functions can be separated into two disjoint subsets $S^{p1}$ and $S^{p2}$; one with partial evaluations computed using variables in $\mathbf{x}^{p1}$ and one using variables in $\mathbf{x}^{p2}$. 
When, the variables inside the $j$-th connected component are inherited from parent $\mathbf{x}$, the partial evaluation is given by: 
\vspace{-2mm}
\begin{equation} 
\label{eq:partial eval}
h_{S_j}(\mathbf{x}) = \sum_{i \in S_j} f_{i}(\mathbf{x})
\end{equation}
For minimization, the best solution, out of $2^{q}$ possible offspring, can be obtained by selecting the components, $S_j$, with best evaluation, i.e., the $i$-th decision variable of the offspring is:
\vspace{-2mm}
\begin{equation} 
\label{eq:z}
x'_{i\in S_j}=
~\left\{ 
\begin{array}{ll}
x^{p1}_i, & \textrm{if $h_{S_j}(\mathbf{x}^{p1}) < h_{S_j}(\mathbf{x}^{p2})$ } \\
x^{p2}_i, & \textrm{if $h_{S_j}(\mathbf{x}^{p2}) \geq h_{S_j}(\mathbf{x}^{p2})$ } 
\end{array} \right.
\end{equation}
 
Here, PX is used in clustering. 
Thus, the integer codification is used instead of binary codification, used in K-bounded pseudo-Boolean optimization. 
This does not prevent the use of PX. 
Since the NKCV2 function is similar to the function used in the NK landscape model, PX can be used without adaptation. 
Here, the time and memory complexity for PX is $O(NK)$, which is $O(N)$ if $K$ is a constant.
PX for clustering is described in Algorithm \ref{alg:PX}\footnote{Figure S1 in the supplementary materials shows an example of recombination by PX.}. 

PX can only be applied when the evaluation of the offspring can be computed as a sum of partial evaluations found in the parents (Eq. (\ref{eq:f_p_xy})). 
In other words, the evaluation function should be decomposable. 
In order to be decomposable, the evaluation function should be a linear sum of subfunctions, each one depending on a small number of objects $K<<N$ (see Eq.~(\ref{eq:NKeval_f})). 
To the best of the authors' knowledge, other internal clustering criteria previously described in the literature are not decomposable. 
As a consequence, PX can be applied only when the NKCV2 criterion is used to evaluate the candidate solutions. 

\begin{algorithm}[ht]
\footnotesize 
\caption{$\mathbf{x}'$=PX\big($\mathbf{x}^{p1}$,$\mathbf{x}^{p2}$\big)}
\label{alg:PX}
\begin{algorithmic}[1]
\STATE $G_{rec}=G_{ep}$
\FOR {$i$=1 \TO $N$}
\IF {$x^{p1}_i = x^{p2}_i$}
\STATE Delete vertex $v_i$ with all its edges from $G_{rec}$
\ENDIF
\ENDFOR
\STATE Use breadth first search to find the $q$ connected components of $G_{rec}$
\FOR {$j$=1 \TO $q$}
\STATE Compute the partial evaluations of the $j$-th connected component for both parents ($h_{j}(\mathbf{x}^{p1})$ and $h_{j}(\mathbf{x}^{p2})$)
\IF {$h_{j}(\mathbf{x}^{p1}) < h_{j}(\mathbf{x}^{p2})$}
\STATE $x'_i=x^{p1}_i$ for all vertices $v_i$ in the $j$-th connected component
\ELSE
\STATE $x'_i=x^{p2}_i$ for all vertices $v_i$ in the $j$-th connected component
\ENDIF
\ENDFOR
\STATE $x'_i=x^{p1}_i$ for all remaining elements
\end{algorithmic}
\end{algorithm}

\section{Experiments}
\label{sec:exp}
Experiments with synthetic and real-world datasets (Section~\ref{sec:exp_des}) are presented. 
First, the NKCV2 criterion is compared to other criteria. 
Then, experiments where the NK hybrid GA is compared to other approaches are presented. 
\vspace{-3mm}
\subsection{Experimental Design}
\label{sec:exp_des}
\vspace{-2mm}
\subsubsection{Datasets}
\label{sec:datasets}
Experiments with three different types of datasets are presented. 
The first type of dataset is generated using the \emph{Gaussian model}, presented in  \cite{milligan1981}. 
Different internal validation criteria have been compared using datasets generated from the Gaussian model \cite{milligan1981,jaskowiak2015}. 
In datasets generated from the Gaussian model, $N$ objects are randomly generated from truncated multivariate normal distributions.  
Each one of the $l$ variables of an object is a normally distributed random variable bounded below and above.  
The model generates $N_c$ non-overlapping clusters from $N_c$ truncated normal distributions, each one with different center and standard deviation. 
The boundaries of the clusters in each of the $l$ dimensions do not overlap, except for the first dimension.
The center and the boundary lengths of each cluster are randomly defined. 
For each dimension, the standard deviation of the respective normal distribution is $1/3$ of the boundary length.  
In the experiments for testing the NK hybrid GA, datasets generated from the Gaussian model with noise are presented. 
The noisy objects are randomly generated according to a uniform distribution. 
The percentage of noisy objects is 1\%\footnote{Figures S2-S5 in the supplementary materials show examples of datasets generated from the Gaussian model with $l=2$.}. 
In the experiments for testing NKCV2, results for datasets generated from the Gaussian model with and without noise are presented. 

We present results with different number of clusters ($N_c=\{2,5,8,11\}$) and number of dimensions ($l=\{2,5\}$). 
The number of objects in the dataset is $N=800$ and the size of each cluster differs according to one of three levels of balance. 
In the first level of balance, the clusters have approximately the same size. 
In the second and third levels, the size of the first cluster is, respectively, $0.1 N$ and $0.6 N$. 
The remaining objects are equally distributed among the other clusters. 
For each combination of $l$, $N_c$, and level of balance, 3 random datasets are generated. 
In this way, the number of datasets generated from the Gaussian model is 72 (3 times the number of combinations of 4 sizes of clusters, 2 dimensions, and 3 levels of balance).

Results for experiments with benchmark problems are also presented (Table \ref{tab:datasets_shapeuci}). 
Seven \emph{shape} datasets, available in \cite{uef_2016}, are employed. 
Results for experiments with four datasets from the UCI machine learning repository \cite{lichman2013} are also presented.  
Some observations about the number of clusters in those datasets are necessary. 
First, the number of classes may not be adequate to evaluate the number of clusters in class-labelled datasets \cite{farber2010}, e.g., the four datasets from the UCI machine learning repository employed here. 
For example, it is well-known that the Iris dataset, with 3 classes, contains only two clusters with obvious separation. 
Second, the values of $N_c$ presented in the tables for the shape datasets are those suggested in the respective references\footnote{Figures S6-S12 in the supplementary materials show the distribution of the objects for each shape dataset with the optimal partition suggested in the respective references.}. 
It is important to observe that $N_c$ is not intrinsic to all of the benchmark problems. 
For example, 5 clusters can be distinguished for the dataset \emph{Aggregation} \cite{gionis2007} if we interpret that a cluster can be formed by 2 dense regions connected by close objects. 
Otherwise, 7 clusters can be identified. 
 
\begin{table}[ht]
\setlength\tabcolsep{4pt} 
\centering
\caption{Shape \cite{uef_2016} and UCI machine learning repository \cite{lichman2013} datasets. 
}
\begin{tabular}{c|cccc} \hline \hline
& Problem & $l$ & $N$ & $N_c$  \\ \hline 
& Aggregation  & 2 & 788 & 7 \\ 
& Zahn´s Compound & 2 & 399 & 6 \\  
Shape & Flame  & 2 & 240 & 2 \\  
& A.K. Jain´s Toy Problem & 2 & 373 & 2 \\  
& Path-based 1   & 2 & 300 & 3 \\ 
& R15  & 2 & 600 & 15 \\ 
& Path-based 2 (Spiral)  & 2 & 312 & 3 \\ 
 \hline
& Iris & 4 & 150 & 3 \\ 
UCI  & Glass & 9 & 124 & 6 \\ 
& Ionosphere & 34 & 351 & 2 \\ 
& Ecoli & 7 & 336 & 8 \\ 
 \hline
\end{tabular}
\label{tab:datasets_shapeuci}
\vspace{-3mm}
\end{table}

\subsubsection{Experiments with the NK Clustering Internal Criterion}
In this set of experiments, evaluation functions based on different internal validation criteria are compared using external validation. 
The method for comparing the evaluation functions is based on that presented in \cite{milligan1985}. 
Given a dataset with known structure, the steps are:
\begin{itemize}
\item[i.] Run clustering algorithms with different parameters, obtaining thus different partitions (candidate solutions);
\item[ii.] Evaluate the quality of the candidate solutions using internal validation criteria;
\item[iii.] Identify the best candidate solution according to each internal validation criterion;
\item[iv.] Evaluate, using external validation, the best candidate solutions.
\end{itemize}

Two external validation measures are employed for evaluating the best candidate solutions. 
The first one is the predicted number of clusters, i.e., the number of clusters in the best candidate solution selected according to each internal validation criterion. 
Several datasets are generated from the Gaussian model for each combination of $N_c$ and $l$. 
In this way, for these datasets, the percentage of hits for the predicted number of clusters is presented. 

The second external validation measure is the \emph{adjusted Rand index} (ARI) for the best partition (candidate solution).  
The ARI is a measure of similarity between two partitions, according to the cluster agreements and disagreements for pairs of objects \cite{hubert1985}. 
Given two partitions $A$ and $B$, the ARI is computed based on the number of pairs of objects that: 
i) are in the same cluster in $A$ and in the same cluster in $B$; 
ii) are in different clusters in $A$ and in different clusters in $B$; 
iii) are in the same cluster in $A$ and in different clusters in $B$; 
iv) are in different clusters in $A$ and in the same cluster in $B$. 
When computing the ARI for a candidate solution $\mathbf{x}$, the partition represented by $\mathbf{x}$ is compared to the optimal partition.

Three algorithms are employed to generate the candidate solutions for each dataset: k-means, DBSCAN, and DP clustering algorithm (see Section~\ref{sec:cluster}). 
Candidate solutions are generated by k-means running the algorithm with the following values of $k$: $2 \leq k \leq \sqrt{N}$ \cite{jaskowiak2015}. 
The best result for running the algorithm 20 times for each value of $k$ is recorded. 
The candidate solutions found by DBSCAN are obtained by running the algorithm with 9 different combinations of parameters $\epsilon$ and $minPts$. 
The parameter $minPts$ assumes one of the following values: $\{\tau_1, \tau_2, \tau_3\}=\{3,4,5\}$ if $l=2$, or  $\{\tau_1, \tau_2, \tau_3\}=\{3,l+1,2l\}$ otherwise. 
The values for $\epsilon$ are determined by the k-nearest neighbour graph with $k=\{\tau_1, 5 \tau_2, 10 \tau_3\}$. 
In the original DP algorithm, the number of prototypes ($\hat N_c$) is manually selected. 
Here (see Section~\ref{sec:cluster}), candidate solutions are obtained by running DP with the following values of $\hat N_c$:  $2 \leq k \leq \sqrt{N}$. 
Thus, the number of candidate solutions evaluated is $2\sqrt{N}+7$ ($\sqrt{N}-1$ for k-means, 9 for DBSCAN, and $\sqrt{N}-1$ for the DP clustering algorithm). 

The internal validation criteria, compared in the experiments, are: the proposed NK clustering validation criterion (NKCV2) with $K=2$, $K=3$ and $K=4$ (Section~\ref{sec:nkeval_func}); silhouette width criterion; DBCV (Section~\ref{sec:eval_func}), and the previous version of the NK clustering validation criterion (NKCV) proposed in \cite{tinos2016}.

\subsubsection{Experiments with the NK Hybrid Genetic Algorithm}
First, the NK hybrid GA (Section~\ref{sec:GA}) is compared to another genetic algorithm: the CGA, described in Section~\ref{sec:eval_func}. 
Then, the NK hybrid GA is compared to k-means, DBSCAN, and DP algorithm. 
Algorithms k-means and DBSCAN are two of the most used state-of-art algorithms for clustering. 
DP clustering algorithm is a recent approach that presents very good performance in many problems. 

Here, CGA uses tournament selection with size of the pool equal to 3 and elitism.
The same is used in the NK hybrid GA. 
Also, the size of the population ($size_{pop}$) is equal: 100 individuals. 
The crossover rate for the NK hybrid GA is $p_c=0.6$ and $K=3$ for the NKCV2 criterion used to evaluate the solutions. 
For CGA, $p_c=0.5$ \cite{hruschka2003}. 
Both algorithms are executed for $N/2$ seconds, i.e., the stopping criteria in Algorithm \ref{alg:hga} is the execution time. 
The number of runs is 25, each one with a different initial random population. 
All experiments were executed in a server with 2 processors Intel Xeon E5-2620 v2 (15 MB Cache, 2.10 GHz) and 32 GB of RAM.

The solutions for k-means, DBSCAN, and DP clustering algorithm are generated as described in the last section. 
When testing the internal evaluation criteria, each algorithm was not analysed independently, i.e., all partitions generated by the three algorithms are used to evaluate the internal criteria independently of knowing which algorithm generates each partition. 
For the evaluation of the clustering algorithms, this information is relevant and the tables show the results for each algorithm. 
In a real-world clustering application, we do not know the information about the desired labels, i.e., we cannot apply external validation to select the best partition. 
In this way, here, we use an internal validation criteria to select the best partition for each algorithm. 
We use DBCV to select the best solutions for each algorithm. 
For the NK hybrid GA, NKCV2 is used to evaluate the solutions during the execution of the algorithm and DBCV is used to select the best solution among the best solutions produced in all runs. 
DBCV is used because it presented better performance than the silhouette width criterion in the experiments with different clustering internal criteria (Section~\ref{sec:res_func}). 
Also, it is not fair to use NKCV2 because the NK hybrid GA optimizes this criterion.
It is important to observe that selecting the partitions according to an internal criterion does not result in the best external validation ever (ARI is used for the external validation). 
However, this is more realistic than selecting the partitions according to the best ARI because this information is not available in real-word clustering.

\vspace{-3mm}
\subsection{Experimental Results}
\label{sec:exp_res}
\vspace{-2mm}
\subsubsection{NK Clustering Internal Criterion}
\label{sec:res_func}
Tables \ref{tab:evalf_gaussian_all}-\ref{tab:evalf_shape_ml} present the results for the comparison of the clustering internal validation criteria. 
NKCV2 with $K=3$ and with $K=4$  present the best results for datasets generated from the Gaussian model without noise (Table \ref{tab:evalf_gaussian_all}). 
For datasets generated from the Gaussian model with noise (Table \ref{tab:evalf_gaussian_all}), three criteria present the best results: NKCV2 with $K=3$, DBCV, and silhouette width criterion. 
DBCV presents the best results for the shape datasets (Table \ref{tab:evalf_shape_ml}), followed by NKCV2 with $K=3$; while DBCV presents better ARI in 5 datasets, NKCV2 with $K=3$ presents better results in 3 datasets. 
The results for dataset Path-based 2 (Spiral) are particularly interesting. 
In this dataset, the objects are disposed in spirals with coincident center. 
The silhouette width criterion is unable to identify good partitions, while DBCV and NKCV2 identify the optimal partitions. 
DBCV and NKCV2 are able to identify the best partitions in this dataset because they efficiently explore the density of the objects in the evaluation of the partitions. 
NKCV2 does not present good results for 2 shape datasets (Flame and A.K. Jain´s Toy Problem). 
We will discuss this when the results for the NK hybrid GA are presented in Section~\ref{sec:res_GA}. 
For the UCI machine learning datasets (Table \ref{tab:evalf_shape_ml}), DBCV presents the best ARI for 1 dataset, while NKCV2 with $K=3$ and silhouette width criterion present the best ARI for 2 datasets each. 

When analysing the total number of datasets in each experiment, we observe that NKCV2 with $K=3$ yields good performance: for each experiment, NKCV2 with $K=3$ was the best or second best when we compute the number of times that an internal validation criterion resulted in the best ARI for more datasets.  
NKCV2 presents, for all datasets, better performance than the previous version of the NK clustering validation criterion (NKCV) proposed in \cite{tinos2016} (the results for UCI machine learning and shape datasets are not presented here). 
Unlike the previous version, the NKCV2 criterion: i) can be used in problems with noise; ii) uses the density of the objects when finding the local groups.

\begin{table*}[ht]
\setlength\tabcolsep{4pt} 
\centering
\caption{Evaluation of clustering internal validation criteria for datasets generated from the Gaussian model. 
The criteria are: NKCV2 with $K=2$, $K=3$ and $K=4$; silhouette width criterion (SWC); density-based clustering validation (DBCV); and NKCV \cite{tinos2016} with  $K=3$. 
The mean ARI for the best hit and the percentage of hits for the correct number of clusters (inside parentheses) are presented. 
The best results for the mean ARI are in bold.}
\begin{tabular}{ccc|cccccc} \hline \hline
Noise & $N_c$ & $l$ &  NKCV2, $K=2$ & NKCV2, $K=3$ & NKCV2, $K=4$&  SWC &    DBCV & NKCV, $K=3$ \\ \hline \hline
No & 2 &2 &\textbf{1.00$\pm$0.00 (100)} &\textbf{1.00$\pm$0.00 (100)} &\textbf{1.00$\pm$0.00 (100)} &0.79$\pm$0.41 (78) &\textbf{1.00$\pm$0.00 (100)} & \textbf{1.00$\pm$0.00 (100)}  \\ 
& 2 &5 &\textbf{1.00$\pm$0.00 (100)} &\textbf{1.00$\pm$0.00 (100)} &\textbf{1.00$\pm$0.00 (100)} &0.82$\pm$0.35 (78)  &\textbf{1.00$\pm$0.00 (100)} & \textbf{1.00$\pm$0.00 (100)} \\ 
& 5 &2 &\textbf{0.99$\pm$0.02 (100)} &\textbf{0.99$\pm$0.02 (100)} &\textbf{0.99$\pm$0.02 (100)} &0.97$\pm$0.08 (89)  &0.83$\pm$0.10 (0) &0.82$\pm$0.19 (22) \\ 
& 5 &5 &\textbf{1.00$\pm$0.00 (100)} &\textbf{1.00$\pm$0.00 (100)} &\textbf{1.00$\pm$0.00 (100)} &0.99$\pm$0.03 (89)  &\textbf{1.00$\pm$0.01 (89)} &\textbf{1.00$\pm$0.01 (89)} \\ 
& 8 &2 &0.87$\pm$0.09 (11) &\textbf{0.89$\pm$0.13 (33)} &\textbf{0.89$\pm$0.13 (33)} &0.88$\pm$0.15 (33)  &0.63$\pm$0.27 (0) &0.56$\pm$0.26 (0) \\ 
& 8 &5 &\textbf{1.00$\pm$0.00 (100)} &\textbf{1.00$\pm$0.00 (100)} &\textbf{1.00$\pm$0.00 (100)} &0.93$\pm$0.13 (67)  &\textbf{1.00$\pm$0.00 (100)} & \textbf{1.00$\pm$0.00 (100)}\\ 
& 11 &2 &0.95$\pm$0.06 (33) &\textbf{0.96$\pm$0.05 (56)} &\textbf{0.96$\pm$0.05 (56)} &0.92$\pm$0.14 (56)  &\textbf{0.96$\pm$0.06 (56)} &0.63$\pm$0.29 (0)  \\ 
& 11 &5 &\textbf{1.00$\pm$0.00 (100)} &\textbf{1.00$\pm$0.00 (100)} &\textbf{1.00$\pm$0.00 (100)} &0.95$\pm$0.14 (89)  &\textbf{1.00$\pm$0.00 (100)} &0.98$\pm$0.06 (78)  \\ 
 \hline
Yes & 2 &2 &0.97$\pm$0.01 (100) &\textbf{0.98$\pm$0.02 (100)} &0.86$\pm$0.32 (89) &0.77$\pm$0.40 (78)  &0.97$\pm$0.01 (100)  &0.76$\pm$0.43 (78) \\ 
 &2 &5 &0.87$\pm$0.32 (89) &0.76$\pm$0.42 (78) &0.76$\pm$0.42 (78) &0.81$\pm$0.34 (78)  &\textbf{0.97$\pm$0.01 (100)} &0.87$\pm$0.32 (89)\\ 
 &5 &2 &0.89$\pm$0.12 (56) &0.86$\pm$0.12 (33) &0.83$\pm$0.11 (11) &\textbf{0.95$\pm$0.08 (78)}  &0.82$\pm$0.10 (0) & 0.76$\pm$0.27 (22)\\ 
 &5 &5 &0.94$\pm$0.10 (67) &0.84$\pm$0.24 (44) &0.81$\pm$0.23 (44) &\textbf{0.98$\pm$0.03 (89)}  &\textbf{0.98$\pm$0.01 (89)} &0.93$\pm$0.10 (56) \\ 
 &8 &2 &0.80$\pm$0.14 (11) &0.75$\pm$0.16 (11) &0.76$\pm$0.16 (11) &\textbf{0.85$\pm$0.16 (33})  &0.60$\pm$0.28 (0) &0.40$\pm$0.24 (0) \\ 
 &8 &5 &\textbf{1.00$\pm$0.00 (100)} &\textbf{1.00$\pm$0.00 (100)} &\textbf{1.00$\pm$0.00 (100)} &0.92$\pm$0.13 (56)  &0.99$\pm$0.00 (100) &0.92$\pm$0.16 (67) \\ 
 &11 &2 &\textbf{0.95$\pm$0.06 (44)} &\textbf{0.95$\pm$0.06 (44)} &\textbf{0.95$\pm$0.06 (44)} &0.89$\pm$0.14 (33)  &0.94$\pm$0.06 (44) &0.71$\pm$0.24 (0) \\ 
 &11 &5 &0.98$\pm$0.04 (78) &0.98$\pm$0.04 (78) &0.98$\pm$0.04 (78) &0.94$\pm$0.14 (78)  &\textbf{0.99$\pm$0.03 (78)} &0.90$\pm$0.11 (44) \\ 
 \hline
\end{tabular}
\label{tab:evalf_gaussian_all}
\vspace{-3mm}
\end{table*}

\subsubsection{NK Hybrid Genetic Algorithm}
\label{sec:res_GA}
Tables \ref{tab:CGA_gaussian}-\ref{tab:CGA_shape_ml} show the results of the experiments comparing CGA and NK hybrid GA. 
The NK hybrid GA presents better performance in all experiments with datasets generated from the Gaussian model with noise. 
The performance is better for 5 out of 7 shape datasets, and in 3 out of 4 UCI machine learning datasets.
For the shape datasets, the NK hybrid GA presents worse results for instances Flame and A.K. Jain´s Toy Problem; NKCV2 presented bad performance in the experiments comparing the internal evaluation criteria for these instances too (Section~\ref{sec:res_func}). 

CGA and NK hybrid GA use different evaluation criteria. 
However, the better performance of the NK hybrid GA is not only due to the use of a different evaluation criterion. 
When comparing the results obtained by CGA (second and third columns in tables \ref{tab:CGA_gaussian}-\ref{tab:CGA_shape_ml}) with the results obtained by k-means, DBSCAN, and DP, selected according to the best silhouette width criterion (tables \ref{tab:evalf_gaussian_all}-\ref{tab:evalf_shape_ml}), it is clear that the performance of CGA is worse than the performance of the best of the three algorithms many times. 
The NK hybrid GA obtains better or similar results (a comparison of NK hybrid GA with k-means, DBSCAN, and DP is made in the second part of this section), which is due to the efficient use of transformation operators and local search mechanism that explicitly use information about the problem structure available in the evaluation function. 

Tables \ref{tab:GA_f9ari_gaussian}-\ref{tab:GA_f9ari_shape_ml} present the results for the comparison of k-means, DBSCAN, DP, and the NK hybrid GA. 
In the tables, the ARI for the best partitions found by each algorithm is presented. 
One can remember that each algorithm was executed several times with different parameters (in the case of the NK hybrid GA, with different random seeds). 
The results shown in the tables are from the best partition found according to DBCV for each algorithm. 
It is important to observe that the best partition found according to one validation criterion is not necessarily the partition with the best ARI among the partitions generated by the algorithm.

The NK hybrid GA presents the best average ARI for the datasets generated according to the Gaussian model with noise. 
When the evaluation (DBCV) is analysed\footnote{Tables S1-S3 in the supplementary materials show the evaluation (DBCV) for the best partitions.}, the best results are for k-means and the NK hybrid GA. 
However, the NK hybrid GA is capable of detecting noise, and k-means, in the standard form employed here, cannot. 
As a consequence, the average ARI is better for the NK hybrid GA in Table \ref{tab:GA_f9ari_gaussian}.

For the shape datasets, DP presents the best performance, with 4 best results, followed by the NK hybrid GA, with 3 best results (Table \ref{tab:GA_f9ari_shape_ml}). 
When DP is better, the NK hybrid GA presents competitive results, with the exception for instance Flame. 
In this problem, the NK evaluation function did not yield good results (Table \ref{tab:evalf_shape_ml}). 
This problem highlights a disadvantage of NKCV2: when there is superposition of two or more clusters, vertices of the interaction graph corresponding to objects located in different clusters are connected by short distance edges. 
As a result, the corresponding clusters are merged. 
In the case of the Flame dataset, almost all objects are close to other objects, i.e., are connected by dense regions. 
As a consequence, the NK hybrid GA finds partitions where practically all objects are located in only one cluster. 
This observation helps to explain the merging of some clusters in some instances generated by the Gaussian model with noise\footnote{The best partitions found for each instance of the shape datasets, and also for some instances generated by the Gaussian model with noise, are illustrated in Figures S2-S12 in the supplementary materials. The optimal solutions and the interaction graph utilized on the methods proposed here are also presented.}. 
For the UCI machine learning datasets (Table \ref{tab:GA_f9ari_shape_ml}), k-means presents the best performance in 3 out of 4 instances. 
DBSCAN presented the best performance for the Ionosphere dataset. 

DBCV was used to select the best solutions generated for every algorithm, including the NK hybrid GA. 
The NK hybrid GA does not employ DBCV during the optimization. 
However, even using a different criterion for selecting the best partitions, the NK hybrid GA produced good performance because it efficiently employs the transformation operators and the local search procedure. 
If NKCV2 is also employed for selecting the best partitions for all algorithms, the results for ARI are similar, despite the fact that the NK hybrid GA presents the best evaluations (NKCV2) when compared with the other algorithms for almost all instances\footnote{Tables S4-S9 in the supplementary materials.}.

As observed in Section~\ref{sec:GA}, the computational complexity of some methods proposed here depends on the maximum outdegree, $K_{out}$, of the interaction graph.  
In the experiments, we observed\footnote{Table S10 in the supplementary materials shows $K_{out}$ for the datasets used here.} that $K_{out}$ is small, when compared to $N$. 
It is also possible to observe that there is a correlation between $K_{out}$ and $l$, but not between $K_{out}$ and $N$ or $N_c$. 

In order to test the impact of the size of the datasets and the dimensionality on the performance of the algorithms, additional experiments were executed. 
First, the clustering algorithms were applied in two datasets from the UCI machine learning repository \cite{lichman2013}: Pen-Based Recognition of Handwritten Digits (Pendigits) and Libras Movement (Librasm). 
While the first dataset has many objects ($N=10,992$), the second dataset has many attributes ($l=90$). 
Also, the algorithms were applied in instance D31, a large shape dataset with 3,100 objects \cite{uef_2016}. 
Second, the clustering algorithms were applied in instances generated from the Gaussian model with fixed number of attributes ($l=10$) and different number of objects ($N$): 2,000; 4,000; 6,000; 8,000; 10,000. 
Finally, the algorithms were applied in instances generated from the Gaussian model with fixed number of objects ($N=1,000$) and different number of attributes ($l$): 100; 300; 500; 700; 900. 
For the instances generated from the Gaussian model, the number of clusters is $N_c=10$. All clusters have approximately the same size (first level of balance). 

In the experiments with large datasets, the NK hybrid GA was executed for $3 N/2$ seconds, except for dataset Pendigits where the algorithm was executed for $4 N$ seconds. 
The best results for datasets Pendigits, Librasm, and D31 were obtained by k-means, followed by the NK hybrid GA (Table \ref{tab:GA_f9ari_large}). 
The NK hybrid GA is generally worse than k-means when there is superposition of spherical clusters or when spherical clusters are very close to each other. 
For dataset D31, k-means found 33 clusters (the optimal solution is composed of 31 clusters). 
However, the NK hybrid GA found only 7 clusters because some of spherical clusters were merged\footnote{Figure S13 in the supplementary materials.}.

For the large Gaussian model datasets, k-means, DBSCAN and the NK hybrid GA obtained the best results (Table \ref{tab:GA_f9ari_large}). 
The ARI for the Gaussian model datasets with large number of objects for the algorithms were even better than the ARI for the datasets with $N=800$ (Table~\ref{tab:GA_f9ari_gaussian}). 
Increasing the size of the search space (by increasing $l$) results in clusters separated by longer distances. 
Increasing the number of objects in the dataset results in clusters with higher density. 
These two properties result in well-defined clusters, making easier the partitioning of the datasets. 
Similar results\footnote{Table S11 in the supplementary materials.} were obtained for the G2 datasets \cite{uef_2016}. 

\begin{table}[ht]
\setlength\tabcolsep{2pt} 
\centering
\caption{Evaluation of clustering internal validation criteria for shape and UCI machine learning datasets. The ARI and the number of clusters (inside parentheses) for the best hit are presented.}
\begin{tabular}{cc|ccccc} \hline \hline
  &  &  & NKCV2 & &   &    \\
  Problem & $N_c$  &  $K=2$ &  $K=3$ & $K=4$&  SWC &   DBCV  \\ \hline \hline
 Aggregation &7 &0.49 (4) &0.49 (4) &0.49 (4) &0.67 (3) &\textbf{0.85 (7)} \\ 
 Zahn´s Compound &6 &\textbf{0.59 (3)} &\textbf{0.59 (3)} &\textbf{0.59 (3)} &0.44 (2) &0.44 (2) \\ 
 Flame &2 &0.00 (1) &0.00 (1) &0.00 (1) &-0.00 (2) &\textbf{0.53 (4)} \\ 
 A.K. Jain´s Toy Prob. &2 &0.00 (1) &0.00 (1) &0.00 (1) &-0.00 (2) &\textbf{0.23 (17)} \\ 
 Path-based1 &3 &0.45 (4) &\textbf{0.47 (3)} &0.43 (2) &0.43 (2)  &0.43 (2) \\ 
 R15 &15 &0.20 (7) &0.20 (7) &0.20 (7) &\textbf{0.97 (15)}  &\textbf{0.97 (15)} \\ 
 Path-based 2 (Spiral) &3 &\textbf{1.00 (3)} &\textbf{1.00 (3)} &\textbf{1.00 (3)} &0.00 (2) &\textbf{1.00 (3)} \\ 
  \hline
 Iris &3 &\textbf{0.57 (2)} &\textbf{0.57 (2)} &\textbf{0.57 (2)} &0.54 (2)  &\textbf{0.57 (2)} \\ 
 Glass &7 &0.21 (5) &\textbf{0.25 (4)} &\textbf{0.25 (4)} &0.19 (2) &0.20 (2) \\ 
 Ionosphere &2 &0.00 (1) &0.00 (1) &0.00 (1) &\textbf{0.18 (2)}  &0.17 (7) \\ 
 Ecoli &8 &0.38 (2) &0.38 (2) &0.38 (2) &\textbf{0.67 (3)} &0.35 (2) \\ 
 \hline
\end{tabular}
\label{tab:evalf_shape_ml}
\vspace{-3mm}
\end{table}

\begin{table}[ht]
\setlength\tabcolsep{4pt} 
\centering
\caption{Results (ARI) for the NK hybrid GA and the CGA for datasets generated from the Gaussian model with noise. 
Columns `mean' present the average results for all runs and all datasets for each combination of $N_c$ and $l$. 
Columns `best' present the average results, over 9 datasets (3 random datasets for each one of the 3 levels of balance) for each combination of $N_c$ and $l$, for the best partitions found by the algorithm (over 25 runs). 
The symbol `s' indicates statistical difference according to the Wilcoxon signed rank test with 0.05 significance level. 
When the test indicated statistical difference, the best results for the mean ARI are in bold. }
\begin{tabular}{cc|cccc} \hline \hline
&  &   \multicolumn{2}{c}{CGA}  &   \multicolumn{2}{c}{NK hybrid GA} \\ 
$N_c$ & $l$ &  mean & best &  mean & best \\ \hline \hline
2 &2 &0.76 $\pm$0.37 &0.78 $\pm$0.35 & \textbf{0.96 $\pm$0.03} (s) &0.97 $\pm$0.03\\ 
2 &5 &0.78 $\pm$0.36 &0.82 $\pm$0.31 &\textbf{0.97 $\pm$0.04} (s) &0.97 $\pm$0.04\\ 
5 &2 &0.76 $\pm$0.06 &0.83 $\pm$0.10 &\textbf{0.99 $\pm$0.00} (s) &0.99 $\pm$0.00\\ 
5 &5 &0.65 $\pm$0.16 &0.84 $\pm$0.19 &\textbf{1.00 $\pm$0.00} (s) &1.00 $\pm$0.00\\ 
8 &2 &0.48 $\pm$0.21 &0.60 $\pm$0.26 &\textbf{0.90 $\pm$0.10} (s) &0.90 $\pm$0.11\\ 
8 &5 &0.45 $\pm$0.20 &0.54 $\pm$0.17 &\textbf{1.00 $\pm$0.00} (s) &1.00 $\pm$0.00\\ 
11 &2 &0.37 $\pm$0.24 &0.47 $\pm$0.21 &\textbf{0.95 $\pm$0.05} (s) &0.95 $\pm$0.05\\ 
11 &5 &0.36 $\pm$0.22 &0.45 $\pm$0.20 &\textbf{0.99 $\pm$0.03} (s) &0.99 $\pm$0.03\\ 
 \hline
\end{tabular}
\label{tab:CGA_gaussian}
\vspace{-3mm}
\end{table}

\begin{table}[ht]
\setlength\tabcolsep{4pt} 
\centering
\caption{Results (ARI) for the NK hybrid GA and the CGA for shape and  UCI machine learning datasets.}
\begin{tabular}{c|cccc} \hline \hline
 &   \multicolumn{2}{c}{CGA}  &   \multicolumn{2}{c}{NK hybrid GA} \\ 
Problem &  mean & best &  mean & best \\ \hline \hline
Aggregation &0.67 $\pm$0.10 &0.70 &\textbf{0.99 $\pm$0.00}  (s) &0.99\\ 
Zahn´s Compound &0.44 $\pm$0.00 &0.44 &\textbf{0.85 $\pm$0.04}  (s) &0.89\\ 
Flame &\textbf{0.51 $\pm$0.04} &0.55 &0.01 $\pm$0.00  (s) &0.01\\ 
A.K. Jain´s Toy Prob. &\textbf{0.30 $\pm$0.00} &0.30 &0.18 $\pm$0.01  (s) &0.19\\ 
Path-based1 &0.40 $\pm$0.00 &0.40 &\textbf{0.50 $\pm$0.03}  (s) &0.57\\ 
R15 &0.13 $\pm$0.04 &0.22 &\textbf{0.87 $\pm$0.01}  (s) &0.87\\ 
Path-based 2 (Spiral) &-0.00 $\pm$0.00 &-0.00 &\textbf{0.89 $\pm$0.18}  (s) &0.99\\ 
\hline
Iris &0.50 $\pm$0.01 &0.50 &\textbf{0.55 $\pm$0.00}  (s) &0.55\\ 
Glass &0.20 $\pm$0.03 &0.23 &0.18 $\pm$0.01 &0.24 \\ 
Ionosphere &0.14 $\pm$0.03 &0.18 &\textbf{0.32 $\pm$0.00}  (s) &0.33\\ 
Ecoli &0.41 $\pm$0.06 &0.61 &\textbf{0.47 $\pm$0.00}  (s) &0.47\\ 
 \hline
\end{tabular}
\label{tab:CGA_shape_ml}
\vspace{-3mm}
\end{table}

\begin{table}[ht]
\setlength\tabcolsep{4pt} 
\centering
\caption{ARI for the best solutions found by 4 algorithms for datasets generated from the Gaussian model with noise. 
The best solution for each algorithm is selected according to DBCV. 
The average ARI, over 9 datasets (3 random datasets for each one of the 3 levels of balance) for each combination of $N_c$ and $l$, for the best partitions is presented.}
\begin{tabular}{cc|cccc} \hline \hline
$N_c$ & $l$ & k-means & DBSCAN & DP &   NK hybrid GA \\ \hline \hline
2 &2 &0.67 $\pm$0.46 &0.61 $\pm$0.44 &\textbf{0.97 $\pm$0.01} &\textbf{0.97 $\pm$0.03} \\ 
2 &5 &\textbf{0.97 $\pm$0.01} &0.29 $\pm$0.32 &\textbf{0.97 $\pm$0.01} &\textbf{0.97 $\pm$0.04} \\ 
5 &2 &0.82 $\pm$0.10 &0.82 $\pm$0.10 &0.79 $\pm$0.27 &\textbf{0.99 $\pm$0.00} \\ 
5 &5 &0.98 $\pm$0.01 &0.81 $\pm$0.23 &0.93 $\pm$0.11 &\textbf{1.00 $\pm$0.00} \\ 
8 &2 &0.65 $\pm$0.25 &0.58 $\pm$0.28 &0.75 $\pm$0.24 &\textbf{0.90 $\pm$0.11} \\ 
8 &5 &0.96 $\pm$0.04 &\textbf{1.00 $\pm$0.00} &0.90 $\pm$0.12 &\textbf{1.00 $\pm$0.00} \\ 
11 &2 &0.94 $\pm$0.05 &\textbf{0.95 $\pm$0.06} &0.90 $\pm$0.14 &\textbf{0.95 $\pm$0.05} \\ 
11 &5 &0.95 $\pm$0.04 &0.98 $\pm$0.04 &0.95 $\pm$0.07 &\textbf{0.99 $\pm$0.03} \\ 
 \hline
\end{tabular}
\label{tab:GA_f9ari_gaussian}
\vspace{-3mm}
\end{table}


\begin{table}[ht]
\setlength\tabcolsep{4pt} 
\centering
\caption{ARI for shape and UCI machine learning datasets.}
\begin{tabular}{c|cccc} \hline \hline
Problem &   k-means & DBSCAN & DP &   NK hybrid GA \\ \hline \hline
Aggregation &0.79 &0.06 &0.85  &\textbf{0.99} \\ 
Zahn´s Compound &0.74 &0.82 &0.44  &\textbf{0.86} \\ 
Flame &0.43 &0.16 &\textbf{0.53}  &0.01 \\ 
A.K. Jain´s Toy Prob. &0.08 &0.09 &\textbf{0.23} &0.17 \\ 
Path-based1 &0.40 &0.50 &0.43  &\textbf{0.57} \\ 
R15 &0.26 &0.26 &\textbf{0.97}  &0.87 \\ 
Path-based 2 (Spiral) &0.12 &0.17 &\textbf{1.00}  &0.99 \\ 
 \hline
Iris &\textbf{0.61} &0.46 &0.57  &0.55 \\ 
Glass &\textbf{0.25} &0.20 &0.17  &0.18 \\ 
Ionosphere &0.11 &\textbf{0.67} &0.17 &0.32 \\ 
Ecoli &\textbf{0.70} &0.18 &0.35  &0.47 \\ 
 \hline
\end{tabular}
\label{tab:GA_f9ari_shape_ml}
\vspace{-3mm}
\end{table}

\begin{table}[ht]
\setlength\tabcolsep{3pt} 
\centering
\caption{ARI for large datasets.}
\begin{tabular}{cccc|cccc} \hline \hline
Problem &  $l$ & $N$ & $N_c$ & k-means & DBSCAN & DP &   NK hybrid GA \\ \hline \hline
Librasm & 90 & 360 & 15 &\textbf{0.31} &0.17 &0.03 &0.19 \\ 
D31 & 2 & 3,100&  31 &\textbf{0.94} &0.00 &0.18  &0.34 \\ 
Pendigits & 16 & 10,992 & 10 &\textbf{0.19} &0.15 &0.14  &0.16 \\ \hline
Gaussian & 10 & 2,000 & 10  &0.99  &\textbf{1.00}  &0.89  &\textbf{1.00}  \\ 
 &  & 4,000 & &0.99 &\textbf{1.00} &0.73 &\textbf{1.00} \\ 
 & & 6,000 &  &\textbf{1.00}  &\textbf{1.00}  &0.10  &\textbf{1.00}  \\ 
 & & 8,000 &  &0.99 &\textbf{1.00} &0.72 &\textbf{1.00} \\ 
 & & 10,000 &  &\textbf{1.00} &\textbf{1.00} &0.89 &\textbf{1.00} \\ \hline
Gaussian & 100 & 1,000 & 10 &  \textbf{1.00} & \textbf{1.00} &0.72  & \textbf{1.00} \\ 
		 & 300 &  & &  \textbf{1.00} & \textbf{1.00} &0.99  & 0.90 \\ 
		 & 500 &  & &  \textbf{1.00} & \textbf{1.00} &0.99  & \textbf{1.00} \\ 
		 & 700 &  &  & \textbf{1.00} & \textbf{1.00} &0.99  & \textbf{1.00} \\ 
	     & 900 &  & &  \textbf{1.00} & \textbf{1.00} &0.89  & \textbf{1.00} \\ 
 \hline 
\end{tabular}
\label{tab:GA_f9ari_large}
\vspace{-3mm}
\end{table}

\section{Conclusions}
\label{sec:con}
The NK hybrid GA\footnote{The source code is freely available in GitHub (https://github.com/rtinos/NKGAclust).} is proposed for hard partitional clustering. 
The NK hybrid GA uses NKCV2 to evaluate the partitions. 
Unlike traditional clustering validation measures, NKCV2 is computed using information about the disposition of small groups of objects. 
Specifically, it is composed of $N$ subfunctions, each one computed using information from a group of $K+1$ objects. 
Objects can belong to different groups, which 
results in a complex network, called interaction graph ($G_{ep}$). 
Experimental results show that density-based regions can be identified by using NKCV2 with fixed small $K$. 
In the NK hybrid GA, after computing $G_{ep}$ and the densities of the objects in the beginning of the run, the time complexity for evaluating each individual using NKCV2 becomes $O(N K)$. 
Experiments showed that a fixed small $K$ is efficient for different numbers of objects ($N$), which reduces the complexity to $O(N)$. 

The information about the relationship between variables allows us to design efficient transformation operators and a local search strategy. 
In other words, we propose the use of gray box optimization in hard partitional clustering problems. 
This is possible because NKCV2 allows one to obtain information about the relationship between variables by inspecting $G_{ep}$. 
Three mutation operators and a local search strategy are proposed, all using the information contained in $G_{ep}$. 
The computational complexity of the mutations and local search strategy depends on the indegree ($K$) and maximum outdegree ($K_{out}$) of $G_{ep}$. 
Experimental results show that $K_{out}$ is small for the clustering instances investigated here. 
Also, small values of $K$ result in good performance. 
NKCV2 allows the use of partition crossover for clustering. 
The evaluation function is decomposed into $q$ independent components by using partition crossover, 
which allows one to deterministically find the best among $2^q$ possible offspring with complexity $O(N K)$, or $O(N)$ if $K$ is a constant. 
In the experiments, the NK hybrid GA produced very good performance when compared to another GA approach and to state-of-art clustering algorithms. 
Experiments also show the efficiency of the NKCV2 criterion, when compared to other criteria. 

In the experiments, the initial population of NK hybrid GA was randomly generated. 
However, solutions from different algorithms, like DBSCAN, k-means, and DP, can be used as initial population. 
Also, the outdegree of $G_{ep}$ was not restricted. 
As $K_{out}$ has an impact in the computational complexity, the investigation of ways of restricting the outdegree of $G_{ep}$ is an important future work. 
NKCV2 is similar to the evaluation function used in the NK landscapes problem. 
Dynamic programming can be efficiently employed in some types of NK landscapes. 
Investigating the adaptation for clustering of dynamic programming, and other state-of-art optimization algorithms used in NK landscapes,  is a relevant future work.


%

\section*{Acknowledgements}
In Brazil, this research was partially funded by FAPESP (2015/06462-1, 2015/50122-0, and 2013/07375-0) and CNPq (303012/2015-3 and 304400/2014-9). 
In Spain, this research was partially funded by Ministerio de Econom\'ia y Competitividad (TIN2014-57341-R and TIN2017-88213-R) and by Ministerio de Educaci\'on Cultura y Deporte (CAS12/00274). 
We would like to thank Dr. Pablo A. Jaskowiak for providing the source code of DBCV.

\ifCLASSOPTIONcaptionsoff
  \newpage
\fi

\bibliographystyle{IEEEtran}

\begin{thebibliography}{10}
\providecommand{\url}[1]{#1}
\csname url@samestyle\endcsname
\providecommand{\newblock}{\relax}
\providecommand{\bibinfo}[2]{#2}
\providecommand{\BIBentrySTDinterwordspacing}{\spaceskip=0pt\relax}
\providecommand{\BIBentryALTinterwordstretchfactor}{4}
\providecommand{\BIBentryALTinterwordspacing}{\spaceskip=\fontdimen2\font plus
\BIBentryALTinterwordstretchfactor\fontdimen3\font minus
  \fontdimen4\font\relax}
\providecommand{\BIBforeignlanguage}[2]{{%
\expandafter\ifx\csname l@#1\endcsname\relax
\typeout{** WARNING: IEEEtran.bst: No hyphenation pattern has been}%
\typeout{** loaded for the language `#1'. Using the pattern for}%
\typeout{** the default language instead.}%
\else
\language=\csname l@#1\endcsname
\fi
#2}}
\providecommand{\BIBdecl}{\relax}
\BIBdecl

\bibitem{halkidi2001}
M.~Halkidi, Y.~Batistakis, and M.~Vazirgiannis, ``On clustering validation
  techniques,'' \emph{Journal of Intelligent Information Systems}, vol.~17,
  no.~2, pp. 107--145, 2001.

\bibitem{kaufman2009}
L.~Kaufman and P.~J. Rousseeuw, \emph{Finding groups in data: an introduction
  to cluster analysis}.\hskip 1em plus 0.5em minus 0.4em\relax John Wiley \&
  Sons, 2009.

\bibitem{jaskowiak2015}
P.~A. Jaskowiak, ``On the evaluation of clustering results: measures,
  ensembles, and gene expression data analysis,'' Ph.D. dissertation,
  University of S\~ao Paulo, November 2015.

\bibitem{macqueen1967}
J.~MacQueen, ``Some methods for classification and analysis of multivariate
  observations,'' in \emph{Proceedings of the 5th Berkeley Symposium on
  Mathematical Statistics and Probability}, 1967, pp. 281--297.

\bibitem{rodriguez2014}
A.~Rodriguez and A.~Laio, ``Clustering by fast search and find of density
  peaks,'' \emph{Science}, vol. 344, no. 6191, pp. 1492--1496, 2014.

\bibitem{falkenauer1998}
E.~Falkenauer, \emph{Genetic algorithms and grouping problems}.\hskip 1em plus
  0.5em minus 0.4em\relax John Wiley \& Sons, Inc., 1998.

\bibitem{hruschka2009}
E.~R. Hruschka, R.~J. G.~B. Campello, A.~Freitas, and A.~C. P. L.~F. Carvalho,
  ``A survey of evolutionary algorithms for clustering,'' \emph{IEEE
  Transactions on Systems, Man, and Cybernetics, Part C: Applications and
  Reviews}, vol.~39, no.~2, pp. 133--155, 2009.

\bibitem{tseng2001}
L.~Y. Tseng and S.~B. Yang, ``A genetic approach to the automatic clustering
  problem,'' \emph{Pattern Recognition}, vol.~34, no.~2, pp. 415--424, 2001.

\bibitem{hruschka2003}
E.~R. Hruschka and N.~F.~F. Ebecken, ``A genetic algorithm for cluster
  analysis,'' \emph{Intelligent Data Analysis}, vol.~7, no.~1, pp. 15--25,
  2003.

\bibitem{tinos2016}
R.~Tin\'{o}s, Z.~Liang, F.~Chicano, and D.~Whitley, ``A new evaluation function
  for clustering: The {NK} internal validation criterion,'' in
  \emph{Proceedings of the 2016 Genetic and Evolutionary Computation
  Conference}, ser. GECCO '16, 2016, pp. 509--516.

\bibitem{moulavi2014}
D.~Moulavi, P.~A. Jaskowiak, R.~J. G.~B. Campello, A.~Zimek, and J.~Sander,
  ``Density-based clustering validation,'' in \emph{Proceedings of the 2014
  SIAM International Conference on Data Mining}, 2014, pp. 839--847.

\bibitem{whitley09}
D.~Whitley, D.~Hains, and A.~Howe, ``Tunneling between optima: partition
  crossover for the {TSP},'' in \emph{Proceedings of the 2009 Conference on
  Genetic and Evolutionary Computation}, ser. GECCO '09, 2009, pp. 915--922.

\bibitem{tinos2015}
R.~Tin\'{o}s, D.~Whitley, and F.~Chicano, ``Partition crossover for
  pseudo-boolean optimization,'' in \emph{Proceedings of the 2015 ACM
  Conference on Foundations of Genetic Algorithms XIII}, ser. FOGA '15, 2015,
  pp. 137--149.

\bibitem{whitley2016}
D.~Whitley, F.~Chicano, and B.~W. Goldman, ``Gray box optimization for {Mk}
  landscapes ({NK} landscapes and {MAX-kSAT}),'' \emph{Evolutionary
  Computation}, vol.~24, no.~3, pp. 491--519, 2016.

\bibitem{chicano2016a}
F.~Chicano, D.~Whitley, and R.~Tin\'{o}s, ``Efficient hill climber for
  multi-objective pseudo-boolean optimization,'' in \emph{European Conference
  on Evolutionary Computation in Combinatorial Optimization}.\hskip 1em plus
  0.5em minus 0.4em\relax Springer, 2016, pp. 88--103.

\bibitem{chicano2016b}
------, ``Efficient hill climber for constrained pseudo-boolean optimization
  problems,'' in \emph{Proceedings of the 2016 Genetic and Evolutionary
  Computation Conference}, ser. GECCO '16, 2016, pp. 309--316.

\bibitem{ester1996}
M.~Ester, H.-P. Kriegel, J.~Sander, and X.~Xu, ``A density-based algorithm for
  discovering clusters in large spatial databases with noise,'' in
  \emph{Proceedings of the 2nd ACM International Conference on Knowledge
  Discovery and Data Mining (KDD)}, 1996, pp. 226--231.

\bibitem{pal1997}
N.~R. Pal and J.~Biswas, ``Cluster validation using graph theoretic concepts,''
  \emph{Pattern Recognition}, vol.~30, no.~6, pp. 847--857, 1997.

\bibitem{yang2004}
J.~Yang and I.~Lee, ``Cluster validity through graph-based boundary analysis.''
  in \emph{Proceedings of the 2004 International Conference on Information and
  Knowledge Engineering (IKE)}, 2004, pp. 204--210.

\bibitem{cheng1995}
Y.~Cheng, ``Mean shift, mode seeking, and clustering,'' \emph{IEEE Transactions
  on Pattern Analysis and Machine Intelligence}, vol.~17, no.~8, pp. 790--799,
  1995.

\bibitem{kauffman93}
S.~A. Kauffman, \emph{The origins of order: Self-organization and selection in
  evolution}.\hskip 1em plus 0.5em minus 0.4em\relax Oxford University Press,
  1993.

\bibitem{milligan1981}
W.~Milligan, ``A monte carlo study of thirty internal criterion measures for
  cluster analysis,'' \emph{Psychometrika}, vol.~46, no.~2, pp. 187--199, 1981.

\bibitem{uef_2016}
\BIBentryALTinterwordspacing
U.~of~Eastern~Finland, ``Clustering datasets,'' accessed: 2016-06-01. [Online].
  Available: \url{https://cs.joensuu.fi/sipu/datasets/}
\BIBentrySTDinterwordspacing

\bibitem{lichman2013}
\BIBentryALTinterwordspacing
M.~Lichman, ``{UCI} machine learning repository,'' 2013, accessed: 2016-06-01.
  [Online]. Available: \url{http://archive.ics.uci.edu/ml}
\BIBentrySTDinterwordspacing

\bibitem{farber2010}
I.~F{\"a}rber, S.~G{\"u}nnemann, H.-P. Kriegel, P.~Kr{\"o}ger, E.~M{\"u}ller,
  E.~Schubert, T.~Seidl, and A.~Zimek, ``On using class-labels in evaluation of
  clusterings,'' in \emph{1st International Workshop on Discovering,
  Summarizing and Using Multiple Clusterings held in conjunction with KDD},
  2010.

\bibitem{gionis2007}
A.~Gionis, H.~Mannila, and P.~Tsaparas, ``Clustering aggregation,'' \emph{ACM
  Transactions on Knowledge Discovery from Data (TKDD)}, vol.~1, no.~1, p.~4,
  2007.

\bibitem{milligan1985}
G.~W. Milligan and M.~C. Cooper, ``An examination of procedures for determining
  the number of clusters in a data set,'' \emph{Psychometrika}, vol.~50, no.~2,
  pp. 159--179, 1985.

\bibitem{hubert1985}
L.~Hubert and P.~Arabie, ``Comparing partitions,'' \emph{Journal of
  Classification}, vol.~2, no.~1, pp. 193--218, 1985.

\end{thebibliography}

%
\vspace{-15mm}
\begin{IEEEbiography}[{\includegraphics[width=1in,height=1.25in,clip,keepaspectratio]{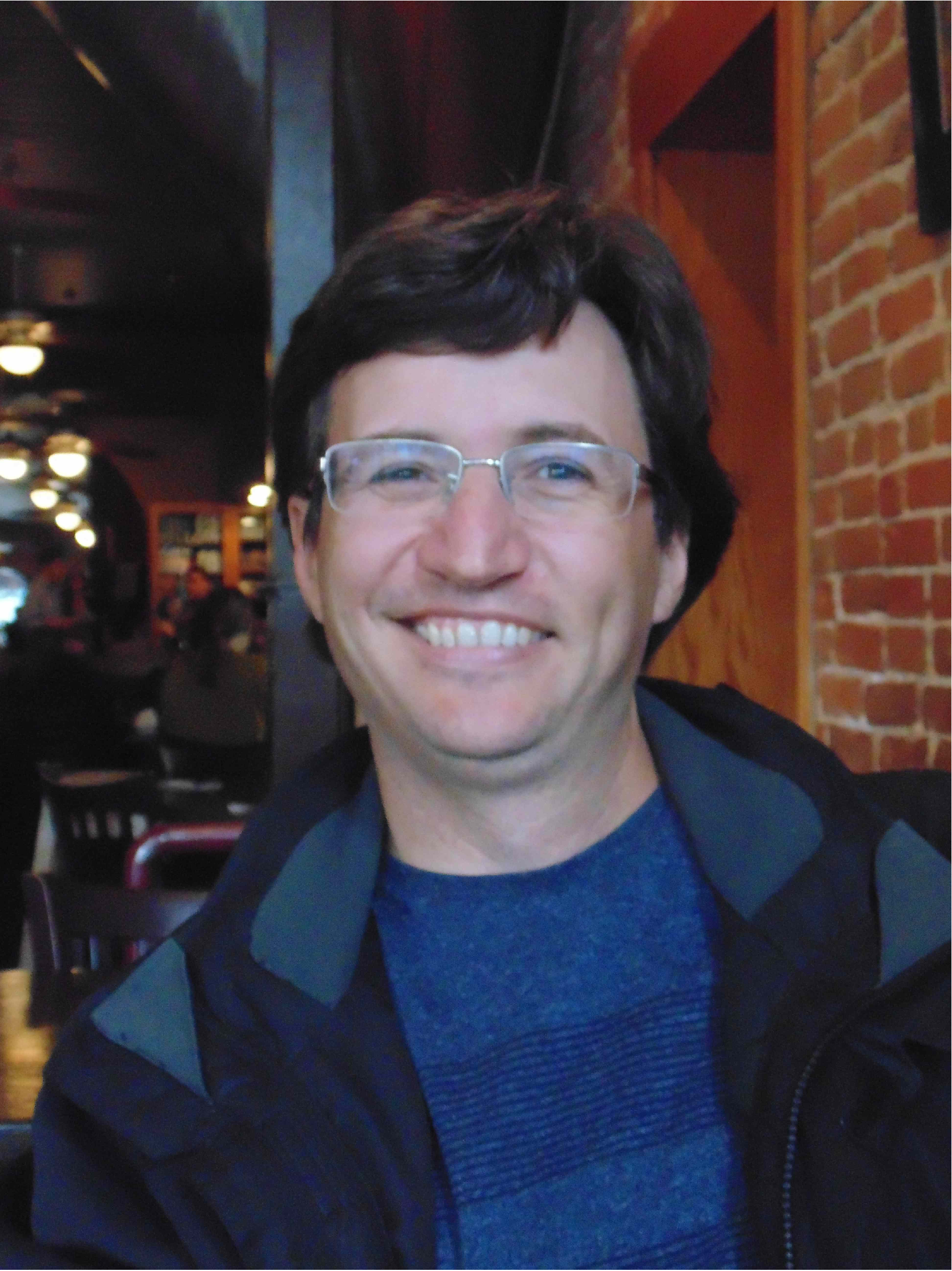}}]{Renato Tin\'os}
received the B.S. degree in electrical engineering from State University of S\~ao Paulo (UNESP), in 1994, and the M.Sc. and
Ph.D. degrees in electrical engineering from the University of S\~ao Paulo (USP) in 1999 and 2003, respectively. 
He joined USP as an Assistant Professor in 2004. 
He is currently an Associate Professor in the Department of Computing and Mathematics of USP at Ribeir\~ao Preto. 
From 2013 to 2014, he was a Visiting Researcher with the Department of Computer Science, Colorado State University.
His research interests include evolutionary algorithms and machine learning. 
\end{IEEEbiography}
\vspace{-15mm}
\begin{IEEEbiography}[{\includegraphics[width=1in,height=1.25in,clip,keepaspectratio]{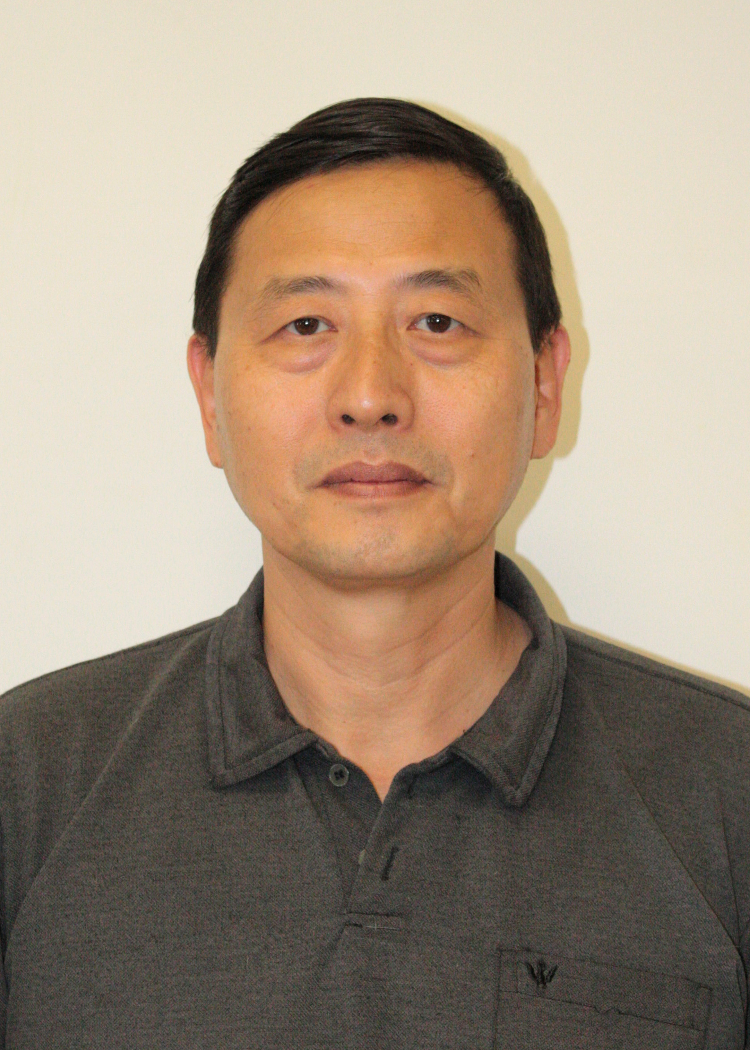}}]{Liang Zhao}
received the B.S. degree from Wuhan University, China, and the 
M.Sc. and Ph.D. degrees from the Aeronautic Institute of Technology, Brazil, in 1988, 1996, 
and 1998, respectively, all in computer science. He joined the University of S\~ao Paulo in 2000, 
where he is currently a Full Professor of the Department of Computing and Mathematics at Ribeir\~ao Preto. 
From 2003 to 2004, he was a Visiting Researcher with the Department of Mathematics, Arizona State 
University. His current research interests include artificial neural networks, machine learning, and complex networks. 
He was an Associate Editor of the IEEE Transactions on Neural Networks and Learning Systems from 2009 to 2012.
\end{IEEEbiography}
\vspace{-15mm}
\begin{IEEEbiography}[{\includegraphics[width=1in,height=1.25in,clip,keepaspectratio]{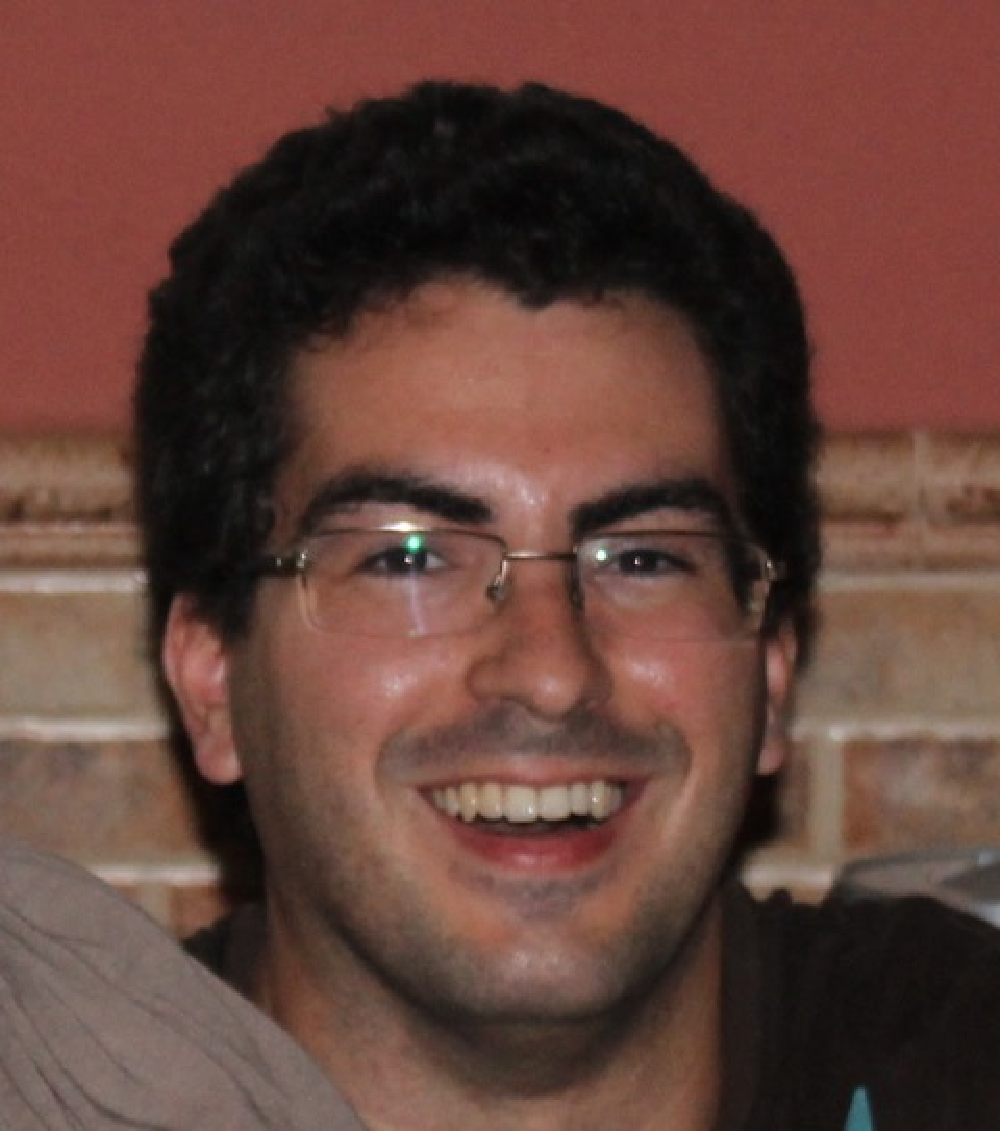}}]{Francisco Chicano}
holds a PhD in Computer Science from the University of M\'alaga and a Degree in Physics from the National Distance Education University. Since 2008 he is with the Department of Languages and Computing Sciences of the University of M\'alaga. His research interests include the application of search techniques to Software Engineering problems and the use of theoretical results to efficiently solve combinatorial optimization problems. He is in the editorial board of Evolutionary Computation Journal, Journal of Systems and Software and Mathematical Problems in Engineering. He has also been programme chair in international events.
\end{IEEEbiography}
\vspace{-15mm}
\begin{IEEEbiography}[{\includegraphics[width=1in,height=1.25in,clip,keepaspectratio]{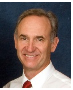}}]{Darrell Whitley}
is Chair of the Computer Science Department at Colorado State University. 
He was editor-in-chief of the journal Evolutionary Computation from 1997 to 2002, and has served on several editorial boards. 
Prof. Whitley served as Chair of the Governing Board of ACM SIGEVO from 2007 to 2011. 
He has been active in Evolutionary Computation since 1986, and has published more than 200 papers with more than 20,000 citations.
\end{IEEEbiography}







\end{document}